\documentclass{article}

\usepackage[preprint,nonatbib]{neurips_2020}

\usepackage[utf8]{inputenc} 
\usepackage[T1]{fontenc}    
\usepackage{hyperref}       
\usepackage{url}            
\usepackage{booktabs}       
\usepackage{amsfonts}       
\usepackage{nicefrac}       
\usepackage{microtype}      

\usepackage{graphicx}
\usepackage{subfigure}
\usepackage{xcolor}

\usepackage{algorithm}
\usepackage{algorithmic}
\usepackage{multirow}
\usepackage{bm}
\usepackage{amsmath}
\usepackage{amssymb}
\usepackage{amsthm}
\newtheorem{theorem}{Theorem}[section]

\newtheorem{proposition}[theorem]{Proposition}

\newtheorem{definition}[theorem]{Definition}
\usepackage{cancel}
\usepackage{caption}
\usepackage{multicol}
\usepackage{threeparttable}

\title{Hierarchical Optimal Transport for \\Robust Multi-View Learning}

\author{%
  Dixin Luo$^{1}$\quad\quad Hongteng Xu$^{1,2}$\quad\quad Lawrence Carin$^{1}$\\
  $^{1}$Department of ECE, Duke University\quad\quad $^{2}$Infinia ML Inc.\\
  \texttt{dixin.luo@duke.edu} \\
}

\begin{document}

\maketitle

\begin{abstract}
Traditional multi-view learning methods often rely on two assumptions: ($i$) the samples in different views are well-aligned, and ($ii$) their representations in latent space obey the same distribution.
Unfortunately, these two assumptions may be questionable in practice, which limits the application of multi-view learning. 
In this work, we propose a hierarchical optimal transport (HOT) method to mitigate the dependency on these two assumptions. 
Given unaligned multi-view data, the HOT method penalizes the sliced Wasserstein distance between the distributions of different views. 
These sliced Wasserstein distances are used as the ground distance to calculate the entropic optimal transport across different views, which  explicitly indicates the clustering structure of the views.
The HOT method is applicable to both unsupervised and semi-supervised learning, and experimental results show that it performs robustly on both synthetic and real-world tasks.
\end{abstract}

\section{Introduction}
Multi-view learning seeks to represent data collected from different sources ($i.e.$, multi-view data) and fuse them in an unsupervised or semi-supervised manner. 
This learning strategy helps fully leverage the information in different views, which is beneficial for many real-world learning tasks, $e.g.$, predicting diseases based on multiple clinical testing records~\cite{yuan2018multi,zhang2018multi} and embedding words semantically across different languages~\cite{guo2012cross}. 
Especially for predictive tasks with few labeled data (and possibly no labels in some views), multi-view learning methods impose useful regularization on target models, and accordingly assist mitigate over-fitting issues.
However, traditional multi-view learning methods are built on two questionable assumptions, which may limit their practical application. 

Firstly, most existing multi-view learning methods~\cite{zhao2017multi,li2018survey} assume that their training data in different views are well-aligned. 
This requirement is inappropriate in many settings. 
For example, real-world multi-view learning often requires us to collect multi-view data from different organizations, $e.g.$, predicting credit level based on account balances from different banks, diagnosing diseases based on clinical and genetic reports from different hospitals, etc.
For security and privacy, the multi-view data from different organizations are anonymous and shuffled before releasing. 
Moreover, it is likely that the views of the data collected by different organizations are generated by different groups of individuals. 
Therefore, real-world multi-view data are often independent, and thus, lack a clear correspondence.
Secondly, the multi-view learning methods based on co-regularization strategies~\cite{andrew2013deep,quang2013unifying,ding2014low,wang2015deep,benton2019deep} assume that the latent representations of the data in different views obey the same distribution. 
In practice, however, the information in a view can be redundant for some views and complementary for the others. 
For such multi-view data, the views have a clustering structure, and thus, obey different latent distributions.
Enforcing a single distribution across all the views may cause serious over-regularization problems. 

\begin{figure}[t]
    \centering
    \includegraphics[width=0.85\linewidth]{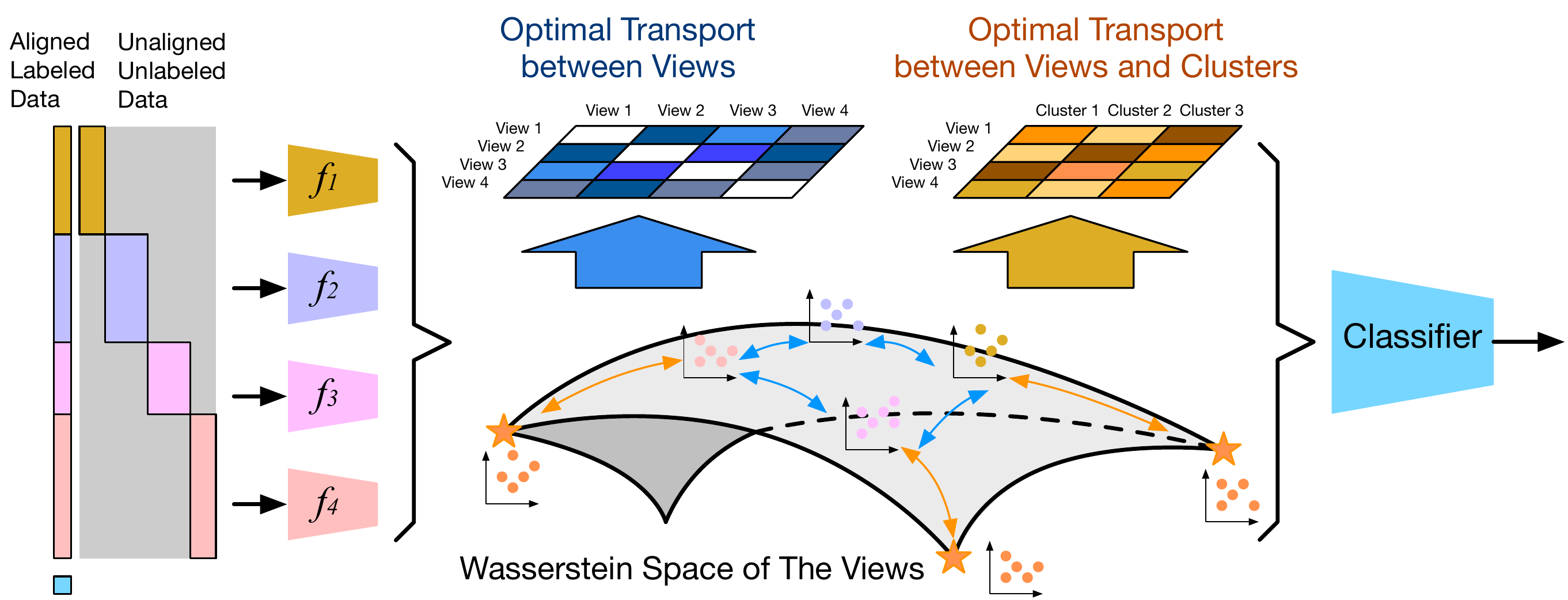}
    \caption{\small{An illustration of our hierarchical optimal transport model for robust multi-view learning. 
    Here, $f_s$ denotes the encoder mapping the samples of the $s$-th view to the latent representation. 
    The classifier takes the concatenation of the latent representations of different views as its input and predicts target labels. 
    The encoders and the classifiers can be trained in a semi-supervised manner when some well-aligned labeled data are available. 
    Based on the sliced Wasserstein distance between the latent representations of different views (the blue arrows) or that between the latent representations and the references (the orange arrows), we build two HOT models and learn optimal transport matrices to indicate the clustering structure of the views.
    }}
    \label{fig:scheme}
\end{figure}

In this work, we propose a new multi-view learning method based on optimal transport theory, mitigating the dependency of multi-view learning on the aforementioned two assumptions.
As illustrated in Figure~\ref{fig:scheme}, for the latent representations of different views, we leverage sliced Wasserstein distance~\cite{bonneel2015sliced} to measure the discrepancy between their distributions, which does not require the correspondence between samples. 
This modification is a generalization of a traditional co-regularization strategy. 
For each pair of views, we introduce a learnable weight to their sliced Wasserstein distance, and these weights can be interpreted as the optimal transport between different views. 
The optimal transport defined on the sliced Wasserstein distances leads to a hierarchical optimal transport (HOT) model. 
It provides a new regularization strategy, that represents the pairwise similarity between different views and implicitly indicates their clustering structure.
Furthermore, given some learnable global representations as references (the orange distributions in Figure~\ref{fig:scheme}), we can apply this HOT model to find the clustering structure of the views explicitly ($i.e.$, the orange optimal transport matrix in Figure~\ref{fig:scheme}).
We learn this HOT model efficiently by combining mini-batch gradient descent~\cite{kingma2014adam} with the Sinkhorn scaling algorithm~\cite{cuturi2013sinkhorn}.
The proposed method achieves robust multi-view learning with fewer assumptions, and is demonstrated to perform well on multiple datasets.

\section{Proposed Model}
Suppose that we have a set of samples collected from $S$ views, $i.e.$, ${X}_s=[{x}_1^s, ..., {x}_N^s]\subset \mathcal{X}_s$ for $s=1,..., S$, where $\mathcal{X}_s$ is the sample space of the $s$-th view, ${x}_n^s\in\mathbb{R}^{D_s}$ for $n=1,...,N$ is a $D_s$-dimensional sample in the space, and ${X}_s$ contains $N$ observed samples. 
We aim to learn $S$ encoders to extract latent representations for the views, and leverage these representations as features for various learning tasks. 
Denote the encoder of the $s$-th view as $f_s:\mathcal{X}_s\mapsto \mathcal{Z}_s$, where $\mathcal{Z}_s$ is the $d_s$-dimensional latent space of view $s$.

Multi-view learning achieves the desired aim by learning the encoders jointly.
We focus on the multi-view learning strategy called \textit{co-regularization}, that leverages the information of one view to impose constraints on the others.  
Typical co-regularization methods include canonical correlation analysis (CCA)~\cite{chaudhuri2009multi} and its variants~\cite{via2007learning,white2012convex,andrew2013deep,ding2014low,wang2015deep,benton2019deep}.
These methods assume that there is a common $d$-dimensional latent space $\mathcal{Z}$ shared by the outputs of the encoders. 
The projections of the encoders' outputs to this space obey the same distribution, or their distributions are highly correlated with each other. 
For example, the Least Squares based Generalized CCA (LSCCA)~\cite{via2007learning} learns the encoders by
\begin{eqnarray}\label{eq:lscca}
\begin{aligned}
\sideset{}{}\min_{\{f_s, {U}_s\}_{s=1}^{S}}  \frac{2}{S(S-1)}\sideset{}{}\sum_{s\neq s'} \|{U}_sf_s({X}_s)-{U}_{s'}f_{s'}({X}_{s'})\|_F^2,~~
\text{s.t.}~\sideset{}{}\sum_{s=1}^{S}{U}_sf_s({X}_s)f_s^{\top}({X}_s){U}_s^{\top} = {I}_d,
\end{aligned}
\end{eqnarray}
where ${U}_s\in\mathbb{R}^{d\times d_s}$ is a matrix projecting the latent representations of each view to a common latent space, and ${I}_d$ is an identity matrix. 
The objective function of LSCCA is the summation of the pairwise comparisons between different views, which penalizes differences between them.
Similarly, we also learn global latent representations shared by all the views, as for Generalized Deep CCA (GDCCA)~\cite{benton2019deep}:
\begin{eqnarray}\label{eq:gdcca}
\begin{aligned}
\sideset{}{_{\{f_s, {U}_s\}_{s=1}^{S}, {G}}}\min  \frac{1}{S}\sideset{}{_{s=1}^{S}}\sum \|{U}_sf_s({X}_s)-{G}\|_F^2,~~
\text{s.t.}~{G}{G}^{\top} = {I}_d,
\end{aligned}
\end{eqnarray}
where ${G}\in\mathbb{R}^{d\times N}$ contains the global latent representations. 
The GDCCA compares different views in an indirect way -- taking ${G}$ as a reference, it makes the latent representations of all the views approach $G$ and suppresses the difference between different views.

\textbf{Semi-supervised Learning.} The above multi-view learning methods can be used as regularizers in semi-supervised learning. 
In particular, when some multi-view data are labeled, we can learn a classifier associated with the encoders by solving the following optimization problem:\footnote{We ignore the constraints associated with $R_M$ to simplify notation.}
\begin{eqnarray}\label{eq:ssl-cca}
\begin{aligned}
\sideset{}{}\min_{\{f_s, {U}_s\}_{s=1}^{S}, g}  \sideset{}{}\sum_{n\in\mathcal{L}} L(g(\{f_s({x}_n^s)\}_{s=1}^{S}), y_n) + \gamma \sideset{}{}\sum_{n\in\mathcal{L}\cup\mathcal{U}} R_M(\{{x}_n^s\}_{s=1}^{S}) +\tau \sideset{}{}\sum_{s=1}^{S}\sideset{}{}\sum_{n\in\mathcal{L}\cup\mathcal{U}} R_S({x}_n^s),
\end{aligned}
\end{eqnarray}
where $\mathcal{L}$ and $\mathcal{U}$ are the sets of indices for labeled and unlabeled data, respectively; 
$y_n$ is the label associated with the $n$-th multi-view data point; and 
$g$ is a classifier taking the concatenation of $\{f_s({x}_n^s)\}_{s=1}^{S}$ as its input. 
The first term in (\ref{eq:ssl-cca}) can be the cross entropy loss for labeled data.
The second term $R_M$ can be an arbitrary regularizer imposed on all the views, which can be implemented as the objective functions in (\ref{eq:lscca}, \ref{eq:gdcca}). 
Finally, the last term $R_S$ can be any additional regularizer imposed on each single view. 
We can implement this term as the manifold-based regularizer that encourages the smoothness of the data manifold~\cite{quang2013unifying,sindhwani2008rkhs}. 
Alternatively, we can introduce $S$ learnable decoders associated with the encoders to construct $S$ autoencoders and implement $R_S$ as the reconstruction loss between the sample in each view and its estimation~\cite{wang2019adversarial,huang2018multimodal,ye2016learning,wang2015deep}.
The two regularizers are weighted by $\gamma$ and $\tau$.

\subsection{Sliced Wasserstein distance for view matching}
The multi-view learning methods in (\ref{eq:lscca}, \ref{eq:gdcca}) require that the samples in different views are well-aligned, $i.e.$, ${x}_n = [{x}_n^1, ..., {x}_n^S]$ for $n=1,..., N$ is sampled jointly from $\mathcal{X}_1\times...\times \mathcal{X}_S$.\footnote{Besides the co-regularization strategy, other multi-view learning strategies like co-training and multi-kernel fusion are also dependent on well-aligned multi-view data, as discussed in Section \ref{sec:related}}
When the samples in each view are generated independently and only a few samples are labeled and well-aligned, as shown in Figure~\ref{fig:scheme}, we need to design a new regularizer to achieve robust multi-view learning without correspondence.
A natural way to modify the objective functions in (\ref{eq:lscca}, \ref{eq:gdcca}) is by introducing permutation matrices to match the samples, $i.e.$, replacing the terms in the objective functions with 
\begin{eqnarray}\label{eq:permute1}
\begin{aligned}
\sideset{}{_{{P}\in\mathcal{P}}}\min\|{U}_sf_s({X}_s){P}-{U}_{s'}f_{s'}({X}_{s'})\|_F^2~\text{for the LSCCA,}
\end{aligned}
\end{eqnarray}
\begin{eqnarray}\label{eq:permute2}
\begin{aligned}
\sideset{}{_{{P}\in\mathcal{P}}}\min\|{U}_sf_s({X}_s){P}-{G})\|_F^2~\text{for the GDCCA.}
\end{aligned}
\end{eqnarray}
where $\mathcal{P}$ represents the set of all valid permutation matrices.\footnote{Without loss of generality, here we assume that for different views, their number of samples are the same. 
Therefore, the ${P}$'s in (\ref{eq:permute1}, \ref{eq:permute2}) are permutation matrices.}
The ${P}$ in (\ref{eq:permute1}) is the permutation matrix indicating the correspondence between the samples of the $s$-th view and those of the $s'$-th view, and the ${P}$ in (\ref{eq:permute2}) is the permutation matrix indicating the correspondence between the samples of the $s$-th view and the global latent representation. 

Because such matching problems are NP-hard, we propose an approximate algorithm to solve them efficiently based on the sliced Wasserstein distance~\cite{bonneel2015sliced,kolouri2018sliced}. 
\begin{definition}[Sliced Wasserstein]
Let $\mathcal{S}^{d-1}=\{{\theta}\in\mathbb{R}^d | \|{\theta}\|_2=1\}$ be the $d$-dimensional hypersphere and $u_{\mathcal{S}^{d-1}}$ the uniform measure on $\mathcal{S}^{d-1}$. 
For each $\theta$, we denote the projection on $\theta$ as $R_\theta$, where $R_{{\theta}}(x)=\langle x, {\theta}\rangle$. 
For arbitrary two probability measures defined on a compact metric space $(\mathcal{X}, d_x)$, denoted as $p$ and $q$, we define their sliced Wasserstein distance as
\begin{eqnarray}\label{eq:sw}
D_{\text{sw}}(p, q) = \mathbb{E}_{\theta\sim u_{\mathcal{S}^{d-1}}}[D_{\text{w}}(R_{\theta}\#p, R_{\theta}\#q)],
\end{eqnarray}
where $R_{\theta}\#p$ is the one-dimensional (1D) distribution after the projection, and $D_{\text{w}}(R_{\theta}\#p, R_{\theta}\#q)$ is the Wasserstein distance between $R_{\theta}\#p$ and $R_{\theta}\#q$ defined on $(R_{\theta}(\mathcal{X}), d_{R_{\theta}(x)})$.
\end{definition}
The sliced Wasserstein distance provides a valid metric to measure the discrepancy between different distributions. 
Given the samples of these two distributions, $i.e.$, ${Z}_1=[{z}_1^1, ..., {z}_N^1]\sim p$ and ${Z}_2=[{z}_1^2, ..., {z}_N^2]\sim q$, we can calculate the sliced Wasserstein distance empirically as
\begin{eqnarray}\label{eq:sw2}
\begin{aligned}
\widehat{D}_{\text{sw}}(Z_1, Z_2) =\frac{1}{M}\sideset{}{}\sum_{m=1}^{M}\sideset{}{}\min_{{P}\in\mathcal{P}}\|\theta_m^{\top}{Z}_1{P}-\theta_m^{\top}{Z}_2\|_2^2
=\frac{1}{M}\sideset{}{}\sum_{m=1}^{M}\|\text{sort}(\theta_m^{\top}{Z}_1)-\text{sort}(\theta_m^{\top}{Z}_2)\|_2^2
\end{aligned}
\end{eqnarray}
where $\{\theta_m\}_{m=1}^{M}$ contains $M$ projectors randomly selected from $\mathcal{S}^{d-1}$, and $\min_{{P}\in\mathcal{P}}\|\theta_m^{\top}{Z}_1{P}-\theta_m^{\top}{Z}_2\|_2^2$ is the empirical estimation of the Wasserstein distance between the two 1D distributions $R_{\theta_m}\#p$ and $R_{\theta_m}\#q$. 
As shown in (\ref{eq:sw2}), when calculating $\widehat{D}_{\text{sw}}(Z_1, Z_2)$, we do not need to learn $M$ permutation matrices explicitly -- for $m=1,...,M$, we just need to sort $\theta_m^{\top}{Z}_1$ and $\theta_m^{\top}{Z}_2$ in ascending (or descending) order,  and calculate the Euclidean distance between the sorted vectors~\cite{kolouri2018sliced}.

Denote the representations of each view's samples in the common latent space ($i.e.$, ${U}_sf_s({X}_s)$) as ${Z}_s$. 
These latent representations can be viewed as the samples of an unknown conditional distribution $P_{\mathcal{Z}|\mathcal{X}_s}$. 
From this standpoint, the matching problems in (\ref{eq:permute1}, \ref{eq:permute2}) empirically measure the discrepancy between different conditional distributions, which can be replaced approximately by the sliced Wasserstein distance in (\ref{eq:sw2}). 
In theory, we have
\begin{proposition}\label{prop1}
Given two sets of samples, denoted as $Z_1$ and $Z_2$, each of which has $N$ $d$-dimensional samples,  $\min_{P\in\mathcal{P}}\|Z_1P-Z_2\|_F^2\geq \widehat{D}_{\text{sw}}(Z_1, Z_2)$.
\end{proposition}
The proof of this proposition is found in the Supplementary Material. 
This result indicates that the sliced Wasserstein distance achieves a lower bound of the optimal objective functions in (\ref{eq:permute1}, \ref{eq:permute2}). 
Therefore, to match the views based on their unaligned samples, we plug the sliced Wasserstein distance into (\ref{eq:lscca}, \ref{eq:gdcca}) and obtain the following two models:
\begin{eqnarray}\label{eq:sw-p}
\begin{aligned}
\sideset{}{}\min_{\{f_s, {U}_s\}_{s=1}^{S}}  \frac{2}{S(S-1)}\sideset{}{}\sum_{s\neq s'} \widehat{D}_{\text{sw}}({U}_sf_s({X}_s),~{U}_{s'}f_{s'}({X}_{s'})),~~
\text{s.t.}~\sideset{}{}\sum_{s=1}^{S}{U}_sf_s({X}_s)f_s^{\top}({X}_s){U}_s^{\top} = {I}_d,
\end{aligned}
\end{eqnarray}
\begin{eqnarray}\label{eq:sw-b}
\begin{aligned}
\sideset{}{_{\{f_s, {U}_s\}_{s=1}^{S}, {G}}}\min  \frac{1}{S}\sideset{}{_{s=1}^{S}}\sum \widehat{D}_{\text{sw}}({U}_sf_s({X}_s),~{G}),~~
\text{s.t.}~{G}{G}^{\top} = {I}_d,
\end{aligned}
\end{eqnarray}

\subsection{Hierarchical optimal transport for view clustering}
The new objective functions in (\ref{eq:sw-p}, \ref{eq:sw-b}) do not require well-aligned samples, but they still tend to make the latent representations of different views approach the same distribution. 
In particular, (\ref{eq:sw-p}) penalizes the sliced Wasserstein distance between each pair of views, while (\ref{eq:sw-b}) penalizes the sliced Wasserstein distance between each view and the reference $G$. 
To overcome this problem, we further modify the multi-view learning methods as follows. 
For (\ref{eq:sw-p}), we introduce learnable weights to the sliced Wasserstein distances and obtain
\begin{eqnarray}\label{eq:sw-hot1}
\begin{aligned}
\sideset{}{}\min_{\{f_s, {U}_s\}_{s=1}^{S}, W} & \sideset{}{}\sum_{s\neq s'} w_{ss'}\widehat{D}_{\text{sw}}({U}_sf_s({X}_s),~{U}_{s'}f_{s'}({X}_{s'})) + \alpha \Bigl\|\sideset{}{}\sum_{s=1}^{S}{U}_sf_s({X}_s)f_s^{\top}({X}_s){U}_s^{\top} - {I}_d\Bigr\|_F^2,\\
\text{s.t.}~&W=W^{\top}\geq 0,~W1_S = \frac{1}{S}1_S,~W^{\top}1_S = \frac{1}{S}1_S,~\text{and}~w_{ss}=0~\text{for}~s=1,..,S.
\end{aligned}
\end{eqnarray}
where $1_S$ represents a $S$-dimensional all-one vector and $W=[w_{ss'}]\in\mathbb{R}^{S\times S}$ is the matrix of the weights.
To avoid trivial solutions ($e.g.$, $W=0$), we restrict $W$ to be ($i$) a doubly stochastic matrix, and ($ii$) a symmetric matrix with all-zero diagonal elements.
By solving this problem, we find the clustering structure of the views implicitly -- the views corresponding to the pairs with large weights belong to the same clusters. 
Note that in (\ref{eq:sw-hot1}) we relax the strict constraint $\sum_{s=1}^{S}{U}_sf_s({X}_s)f_s^{\top}({X}_s){U}_s^{\top} = {I}_d$ to a least squares based regularizer, which helps us to apply mini-batch gradient descent directly to learn the model. 
In the subsequent experiments, we find that this relaxation does not do harm to the learning results while simplifies our learning algorithm.

For (\ref{eq:sw-b}), besides introducing learnable weights, we consider multiple global latent representations which correspond to different clusters directly. 
The problem becomes
\begin{eqnarray}\label{eq:sw-hot2}
\begin{aligned}
\sideset{}{_{\{f_s, {U}_s\}_{s=1}^{S}, \{G_k\}_{k=1}^{K}, W}}\min & \sideset{}{_{s=1}^{S}}\sum\sideset{}{_{k=1}^{K}}\sum w_{sk}\widehat{D}_{\text{sw}}({U}_sf_s({X}_s),{G}_k) + \alpha\Bigl\|\sideset{}{_{k=1}^{K}}\sum {G}_k{G}_k^{\top} - {I}_d\Bigr\|_F^2,\\
\text{s.t.}~&W\geq 0,~W1_K = \frac{1}{S}1_S,~W^{\top}1_S = \frac{1}{K}1_K.
\end{aligned}
\end{eqnarray}
where $K$ is the number of clusters we set for the views, which is fixed as 3 in the following experiments; 
$G_k\in \mathbb{R}^{d\times N}$ represents the global latent representation matrix corresponding to the $k$-th cluster; and  
$W=[w_{sk}]\in\mathbb{R}^{S\times K}$ is the matrix of the weights, which is restricted as a doubly stochastic matrix. 
According to its constraints, we can explain the matrix as the joint distribution of the views and the clusters, and the element $w_{sk}$ is the probability that the $s$-th view belongs to the $k$-th cluster. 
In other words, this method can find the clustering structures explicitly.
Similar to (\ref{eq:sw-hot1}), we relax the strict constraint $\sum_{k=1}^{K} {G}_k{G}_k^{\top} = {I}_d$ to a regularizer in (\ref{eq:sw-hot2}). 

In both these two methods, we establish an optimal transport model with a hierarchical architecture. 
The $W$ in (\ref{eq:sw-hot1}) achieves an optimal transport across different views, whose ground distance is the sliced Wasserstein distance between the latent representations of the views. 
Similarly, the $W$ in (\ref{eq:sw-hot2}) is an optimal transport from the views to their clusters, whose ground distance is the sliced Wasserstein distance between the latent representations of the views to those of the clusters. 
These optimal transport matrices can be learned efficiently by computing the entropic Wasserstein distance based on the Sinkhorn scaling algorithm~\cite{cuturi2013sinkhorn}. 
To our knowledge, our work is the first to leverage the hierarchical optimal transport model to implement multi-view learning methods. 
This framework provides a new way to represent different views and find their clustering structure.

\begin{figure*}[t]
\centering
\begin{minipage}[t]{0.49\linewidth}
\begin{algorithm}[H]
\small{
	\caption{The optimization of (\ref{eq:sw-hot1})}
	\label{alg1}
	\begin{algorithmic}[1]
	    \STATE Initialize $W=\frac{1}{S(S-1)}(1_S1_S^{\top} - I_S)$.
	    \STATE \textbf{For} each epoch:
	    \STATE ~Sample batches $\{X_s\}_{s=1}^{S}$ from the views.
		\STATE ~Calculate $C=[\widehat{D}_{\text{sw}}(U_sf_s(X_s), U_{s'}f_{s'}(X_{s'}))]$.
		\STATE ~Calculate $c=1_S^{\top}C1_S$.
		\STATE ~Learn $W$ by the \textbf{Sinkhorn algorithm}~\cite{cuturi2013sinkhorn}:\\
		       ~$\min_{W\in\Pi(\frac{1}{S},\frac{1}{S})} \langle W, C+cI_S\rangle + \beta\langle W, \log W\rangle$.
		\STATE ~Fix $W$ and calculate the loss function $R_M$ as\\
		       ~$\langle W, C\rangle$+$\alpha\|\sum_{s}{U}_sf_s({X}_s)f_s^{\top}({X}_s){U}_s^{\top} - {I}_d\|_F^2$.
		\STATE ~Update the model by \textbf{Adam}~\cite{kingma2014adam}:\\
		       ~$\{f_s, U_s\}_{s=1}^{S}\leftarrow \text{Adam}(R_M)$.
	\end{algorithmic}
}
\end{algorithm}
\end{minipage}
\vspace{-8pt}
~
\begin{minipage}[t]{0.49\linewidth}
\begin{algorithm}[H]
\small{
	\caption{The optimization of (\ref{eq:sw-hot2})}
	\label{alg2}
	\begin{algorithmic}[1]
	    \STATE Initialize $W=\frac{1}{SK}1_S1_K^{\top}$.
	    \STATE Initialize $\{G_k\}_{k=1}^{K}$ as $K$ random matrices.
	    \STATE \textbf{For} each epoch:
	    \STATE ~Sample batches $\{X_s\}_{s=1}^{S}$ from the views.
		\STATE ~Calculate $C=[\widehat{D}_{\text{sw}}(U_sf_s(X_s), G_k)]$.
		\STATE ~Learn $W$ by the \textbf{Sinkhorn algorithm}~\cite{cuturi2013sinkhorn}:\\
		       ~$\min_{W\in\Pi(\frac{1}{S},\frac{1}{K})} \langle W, C\rangle + \beta\langle W, \log W\rangle$.
		\STATE ~Fix $W$ and calculate the loss function $R_M$ as\\
		       ~$\langle W, C\rangle+\alpha\|\sum_{k}G_k G_k^{\top} - {I}_d\|_F^2$.
		\STATE ~Update the model by \textbf{Adam}~\cite{kingma2014adam}:\\
		       ~$\{f_s, U_s\}_{s=1}^{S}, \{G_k\}_{k=1}^{K}\leftarrow \text{Adam}(R_M)$.
	\end{algorithmic}
}
\end{algorithm}
\end{minipage}
\vspace{-8pt}
\end{figure*}

\section{Learning Algorithm}
We propose an efficient learning algorithm to solve the problems in (\ref{eq:sw-hot1}, \ref{eq:sw-hot2}), based on alternating optimization. In each iteration, we first calculate the sliced Wasserstein distances and update the weight matrix via the Sinkhorn scaling algorithm~\cite{cuturi2013sinkhorn}. 
Then, we fix the weight matrix and learn the encoders and their projection matrices via mini-batch gradient descent, $i.e.$, the Adam algorithm~\cite{kingma2014adam}. 
Algorithms~\ref{alg1} and~\ref{alg2} show the details of our implementation, where $\Pi(p, q)=\{W\geq 0|W1=p,W^{\top}1=q\}$ is the set of  doubly stochastic matrices with marginals $p$ and $q$, and  
$\beta$ is the weight of the entropic regularizer when applying the Sinkhorn algorithm.
The details of the Sinkhorn algorithm can be found in~\cite{cuturi2013sinkhorn} and in our Supplementary Material.
In line 6 of Algorithm~\ref{alg1}, we set the cost matrix to $C+cI_s$, to ensure $w_{ss}=0$ for $s=1,...,S$. 
Additionally, the Sinkhorn algorithm can make $W$ converge to a symmetric matrix when the cost matrix is symmetric and the marginals of $W$ are the same. 
Therefore, all the constraints on $W$ can be readily satisfied.
In Algorithm~\ref{alg2}, the $\{G_k\}_{k=1}^{K}$ in line 2 are initialized as Gaussian random matrices, and for each the number of columns is equal to the batch size. 
Moreover, when some well-aligned labeled data are available, we can apply these two algorithms to achieve semi-supervised learning. 
Considering the labeled data, we just need to replace the loss $R_M$ in Algorithms~\ref{alg1} and \ref{alg2} with the loss $L+R_M+R_S$ in (\ref{eq:ssl-cca}) and update the model and a classifier jointly.

Our HOT model is a new member of the hierarchical optimal transport family, combining sliced Wasserstein distance with entropic Wasserstein distance. 
Compared with existing hierarchical optimal transport models~\cite{chen2018optimal,lee2019hierarchical,yurochkin2019hierarchical,xu2020learning}, our model has advantages from the perspectives of computational complexity and model flexibility. 
Given $S$ views, each of which contains $N$ samples in a batch, the computational complexity of our method is $\mathcal{O}(S^2(MN + J))$ for Algorithm~\ref{alg1} and $\mathcal{O}(SK(MN + J))$ for Algorithm~\ref{alg2}. 
Here, $M$ is the number of random projections used to compute a sliced Wasserstein distance, which is much smaller than $N$, $J$ is the number of iterations used in the Sinkhorn scaling algorithm, and $K$ is the number of clusters for the views. 
The first term $\mathcal{O}(S^2MN)$ ($\mathcal{O}(SKMN)$) corresponds to calculating the cost matrix based on sliced Wasserstein distance,  
and the second term $\mathcal{O}(S^2J)$ ($\mathcal{O}(SKJ)$) corresponds to computing the entropic Wasserstein distance based on the Sinkhorn scaling algorithm. 
Instead of using sliced Wasserstein distance, existing hierarchical optimal transport models apply Wasserstein distance~\cite{chen2018optimal,yurochkin2019hierarchical,xu2020learning} or entropic Wasserstein distance~\cite{lee2019hierarchical} to calculate the cost matrix $C$. 
As a result, for each element of the cost matrix, the computational complexity is $\mathcal{O}(N^3)$ when applying linear programming to compute Wasserstein distance, or $\mathcal{O}(JN^2)$ when applying Sinkhorn scaling algorithm to compute entropic Wasserstein distance~\cite{lee2019hierarchical,yurochkin2019hierarchical}. 
To avoid these computations, such methods have to assume the distribution of the samples to be Gaussian~\cite{chen2018optimal,xu2020learning}, which limits their applicability and increases the risk of over-regularization. 
According to the analysis above, our learning algorithms has much less computational complexity, without the need to impose any assumptions on the latent distributions of the views.

\section{Related Work\label{sec:related}}
\textbf{Multi-view learning} Multi-view learning can be broadly categorized into three strategies~\cite{zhao2017multi,sun2013survey}: co-training, multi-kernel fusion, and co-regularization~\cite{guo2019canonical}. 
Co-training requires ($i$) that the views in the training data are conditionally independent, and ($ii$) that each view is sufficient to predict labels. It iteratively learns a separate classifier for each view using labeled samples and annotates the unlabeled data based on the most confident predictions of each classifier~\cite{kumar2011co,ma2017self}. 
Kernel-based methods merge the kernel matrices of different views and learn global representations based on the merged kernel~\cite{de2010multi,li2015large}. 
Co-regularization methods add regularization terms to encourage the data from different views to be consistent. 
The representative regularizers include ($i$) CCA-based methods~\cite{guo2019canonical,chaudhuri2009multi,sindhwani2008rkhs,via2007learning,guo2012cross,andrew2013deep} that penalize the difference between the views in the latent space, and ($ii$) linear discriminate analysis based methods~\cite{jin2014multi} that require labeled data. 
Generally, the co-training methods are not scalable for cases with more than two views, and the kernel-based methods are transductive. 
Because both approaches are not suitable for large-scale multi-view learning tasks, in our work we focus on the co-regularization strategy and its improvements.
Additionally, all methods discussed above require well-aligned multi-view data. 
Although some methods have been proposed to achieve multi-view learning based on incomplete or noisy views~\cite{xu2015multi,jin2014multi,christoudias2008multi}, they rely on labeled data, which are not available in many scenarios.

\textbf{Optimal transport-based learning} Optimal transport theory~\cite{villani2008optimal} has proven to be useful in distribution matching~\cite{memoli2011gromov,su2017order}, data clustering~\cite{agueh2011barycenters,xu2018distilled,yurochkin2019hierarchical}, and learning a generative model~\cite{arjovsky2017wasserstein}. 
Given two sets of samples, we can calculate the optimal transport between them by linear programming~\cite{kusner2015word}.
With the help of the Sinkhorn algorithm~\cite{benamou2015iterative}, an entropic Wasserstein distance has been proposed to accelerate the computation of optimal transport \cite{cuturi2013sinkhorn}.
When only the distance between distributions is needed, one can apply the dual form of Wasserstein distance~\cite{arjovsky2017wasserstein} or the sliced Wasserstein distance~\cite{bonneel2015sliced,kolouri2018sliced} to approximate the distance, and avoid explicitly computing the optimal transport. 
Recently, hierarchical optimal transport models have been proposed to compare the distributions with structural information, $e.g.$, the nonlinear factorization models in~\cite{xu2018distilled,xu2020gromov,schmitzer2013hierarchical}, and optimal transport models for multi-modal distributions~\cite{chen2018optimal,lee2019hierarchical,yurochkin2019hierarchical}.
These hierarchical optimal transport models achieve encouraging performance on multi-modal distribution matching~\cite{chen2018optimal,lee2019hierarchical,xu2020learning} and data clustering~\cite{yurochkin2019hierarchical,xu2020gromov}. 
Compared with existing HOT models, our model has lower computational complexity, and it does not have constraints on the target distributions.

\begin{table}[t]
\begin{small}
    \caption{The statistics of each real-world dataset}\label{tab:stats}
    \centering
    \begin{tabular}{
    @{\hspace{0pt}}c@{\hspace{1pt}}|
    @{\hspace{1pt}}c@{\hspace{1pt}}|
    @{\hspace{1pt}}c@{\hspace{1pt}}|
    @{\hspace{1pt}}c@{\hspace{1pt}}|
    @{\hspace{1pt}}c@{\hspace{1pt}}|
    @{\hspace{1pt}}c@{\hspace{1pt}}|
    @{\hspace{1pt}}c@{\hspace{1pt}}|
    @{\hspace{1pt}}c@{\hspace{1pt}}|
    @{\hspace{1pt}}c@{\hspace{0pt}}
    }
    \hline\hline
    Dataset
    &\# Samples
    &\# Cls.
    &View 1/$D_1$
    &View 2/$D_2$
    &View 3/$D_3$
    &View 4/$D_4$
    &View 5/$D_5$
    &View 6/$D_6$
    \\ \hline
    Caltech-7/20
    &1474/2386
    &7/20
    &Gabor/48
    &WM/40
    &CENTRIST/254
    &HOG/1984
    &GIST/512
    &LBP/928\\
    Handwritten
    &2000
    &10
    &Pixel/240
    &Fourier/76
    &FAC/216
    &ZER/47
    &KAR/64
    &MOR/6\\
    \hline\hline
    \end{tabular}
\end{small}
\end{table}

\section{Experiments}
To demonstrate the usefulness of the proposed multi-view learning methods, we test them on both synthetic and real-world datasets, with  comparisons to state-of-the-art methods. 
For each method, we consider the following four datasets: Caltech7, Caltech20, and the Handwritten datasets in~\cite{li2015large}.
The Caltech7, Caltech20, and Handwritten datasets correspond to three image classification tasks. 
Each contains 5-6 kinds of visual features extracted by classic methods. 
The details of the feature extraction methods are provided at \url{https://github.com/yeqinglee/mvdata}.
The statistics of these datasets are summarized in Table~\ref{tab:stats}. 
For each dataset, we test our method and the alternative approaches in 20 trials. 
In each trial, we randomly select 60\% of the samples for training, 20\% of the samples for validation, and the remaining 20\% of samples for testing. 
For the training data, we keep 5\% as well-aligned and labeled data. 
For existing multi-view learning methods that require well-aligned data, we just remove the labels of the remaining training data. 
Concerning the proposed robust multi-view learning methods and their variants, we randomly permute the remaining training data in each view and remove their labels to generate unaligned unlabeled data. 

\begin{table}[t]
\begin{small}
\begin{center}
    \caption{Averaged classification accuracy (\%) and standard deviation (semi-supervised learning)}\label{tab:result1}
    \centering
    \begin{tabular}{
    @{\hspace{4pt}}c@{\hspace{4pt}}|
    @{\hspace{4pt}}c@{\hspace{4pt}}
    @{\hspace{4pt}}c@{\hspace{4pt}}
    @{\hspace{4pt}}c@{\hspace{4pt}}|
    @{\hspace{4pt}}c@{\hspace{4pt}}
    @{\hspace{4pt}}c@{\hspace{4pt}}
    @{\hspace{4pt}}c@{\hspace{4pt}}
    }
    \hline\hline
    Method &$R_M$ &$R_S$ &Data &Caltech7 &Caltech20 &Handwritten\\
    \hline
    \multirow{3}{*}{Baseline}&LSCCA~\cite{via2007learning} 	
    &---
    &Aligned
    &87.36$\pm$1.43 	&71.20$\pm$2.74 	&87.98$\pm$3.46\\
    &DGCCA~\cite{benton2019deep} 
    &---
    &Aligned
    &87.60$\pm$1.08 	&71.80$\pm$2.61 	&87.12$\pm$3.89\\
    &AECCA~\cite{wang2015deep} 	
    &AE
    &Aligned
    &87.62$\pm$1.47 	&71.50$\pm$2.84 	&88.53$\pm$3.22\\
    \hline
    \multirow{8}{*}{Ours}&SW~(\ref{eq:sw-p})
    &---
    &Unaligned
    &87.11$\pm$1.55 	&70.94$\pm$3.05 	&88.58$\pm$3.29\\
    &SW~(\ref{eq:sw-b})
    &---
    &Unaligned 	
    &87.55$\pm$1.79 	&72.50$\pm$2.41 	&89.78$\pm$2.98\\
    &HOT~(\ref{eq:sw-hot1})
    &---
    &Unaligned
    &88.31$\pm$1.56 	&72.96$\pm$2.69 	&89.95$\pm$3.52\\
    &HOT~(\ref{eq:sw-hot2}), $K=3$
    &---
    &Unaligned 	
    &88.29$\pm$1.87 	&73.43$\pm$2.70 	&90.05$\pm$2.62\\
    &SW~(\ref{eq:sw-p})
    &AE
    &Unaligned 	 	
    &87.49$\pm$1.25 	&70.94$\pm$2.73 	&89.90$\pm$3.31\\
    &SW~(\ref{eq:sw-b})
    &AE
    &Unaligned 	 	
    &87.24$\pm$1.64 	&72.36$\pm$2.27 	&89.22$\pm$3.28\\
    &HOT~(\ref{eq:sw-hot1})
    &AE
    &Unaligned 		
    &88.47$\pm$1.52 	&73.18$\pm$2.40 	&90.37$\pm$3.00\\
    &HOT~(\ref{eq:sw-hot2}), $K=3$
    &AE
    &Unaligned 		
    &88.98$\pm$2.15 	&73.48$\pm$2.08 	&91.07$\pm$2.55\\
    \hline\hline
    \end{tabular}
\end{center}
\end{small}
\end{table}

\begin{table}[t]
\begin{small}
\begin{center}
    \caption{Averaged classification accuracy (\%) and standard deviation (unsupervised)}\label{tab:result2}
    \centering
    \begin{tabular}{
    @{\hspace{4pt}}c@{\hspace{4pt}}|
    @{\hspace{4pt}}c@{\hspace{4pt}}
    @{\hspace{4pt}}c@{\hspace{4pt}}
    @{\hspace{4pt}}c@{\hspace{4pt}}|
    @{\hspace{4pt}}c@{\hspace{4pt}}
    @{\hspace{4pt}}c@{\hspace{4pt}}
    @{\hspace{4pt}}c@{\hspace{4pt}}
    }
    \hline\hline
    Method &$R_M$ &$R_S$ &Data &Caltech7 &Caltech20 &Handwritten\\
    \hline
    \multirow{3}{*}{Baseline}&LSCCA~\cite{via2007learning} 	
    &---
    &Aligned
    &82.33$\pm$1.84 	&60.42$\pm$2.16 	&70.20$\pm$7.77\\
    &DGCCA~\cite{benton2019deep} 
    &---
    &Aligned
    &75.56$\pm$4.59 	&54.67$\pm$3.13 	&66.15$\pm$5.20\\
    &AECCA~\cite{wang2015deep} 	
    &AE
    &Aligned
    &84.39$\pm$1.71 	&66.75$\pm$2.47 	&83.30$\pm$4.83\\
    \hline
    \multirow{8}{*}{Ours}&SW~(\ref{eq:sw-p})
    &---
    &Unaligned
    &82.25$\pm$1.47 	&59.93$\pm$3.74 	&65.67$\pm$9.80\\
    &SW~(\ref{eq:sw-b})
    &---
    &Unaligned 	
    &85.71$\pm$1.33 	&69.21$\pm$2.51 	&86.40$\pm$2.71\\
    &HOT~(\ref{eq:sw-hot1})
    &---
    &Unaligned
    &82.89$\pm$1.32 	&60.85$\pm$3.31 	&67.65$\pm$7.56\\
    &HOT~(\ref{eq:sw-hot2}), $K=3$
    &---
    &Unaligned 	
    &86.27$\pm$1.79 	&71.07$\pm$2.29 	&87.12$\pm$2.10\\
    &SW~(\ref{eq:sw-p})
    &AE
    &Unaligned 	 	
    &83.88$\pm$1.70 	&67.27$\pm$2.96 	&82.73$\pm$3.70\\
    &SW~(\ref{eq:sw-b})
    &AE
    &Unaligned 	 	
    &86.64$\pm$1.57 	&69.29$\pm$2.37 	&87.63$\pm$3.08\\
    &HOT~(\ref{eq:sw-hot1})
    &AE
    &Unaligned 		
    &83.98$\pm$1.55 	&67.41$\pm$2.32 	&83.85$\pm$3.84\\
    &HOT~(\ref{eq:sw-hot2}), $K=3$
    &AE
    &Unaligned 		
    &87.32$\pm$1.33 	&69.48$\pm$2.53 	&87.50$\pm$3.15\\
    \hline\hline
    \end{tabular}
\end{center}
\end{small}
\end{table}

\begin{figure}[t]
\centering
\begin{minipage}{0.4\textwidth}
\centering
    \includegraphics[height=3.5cm]{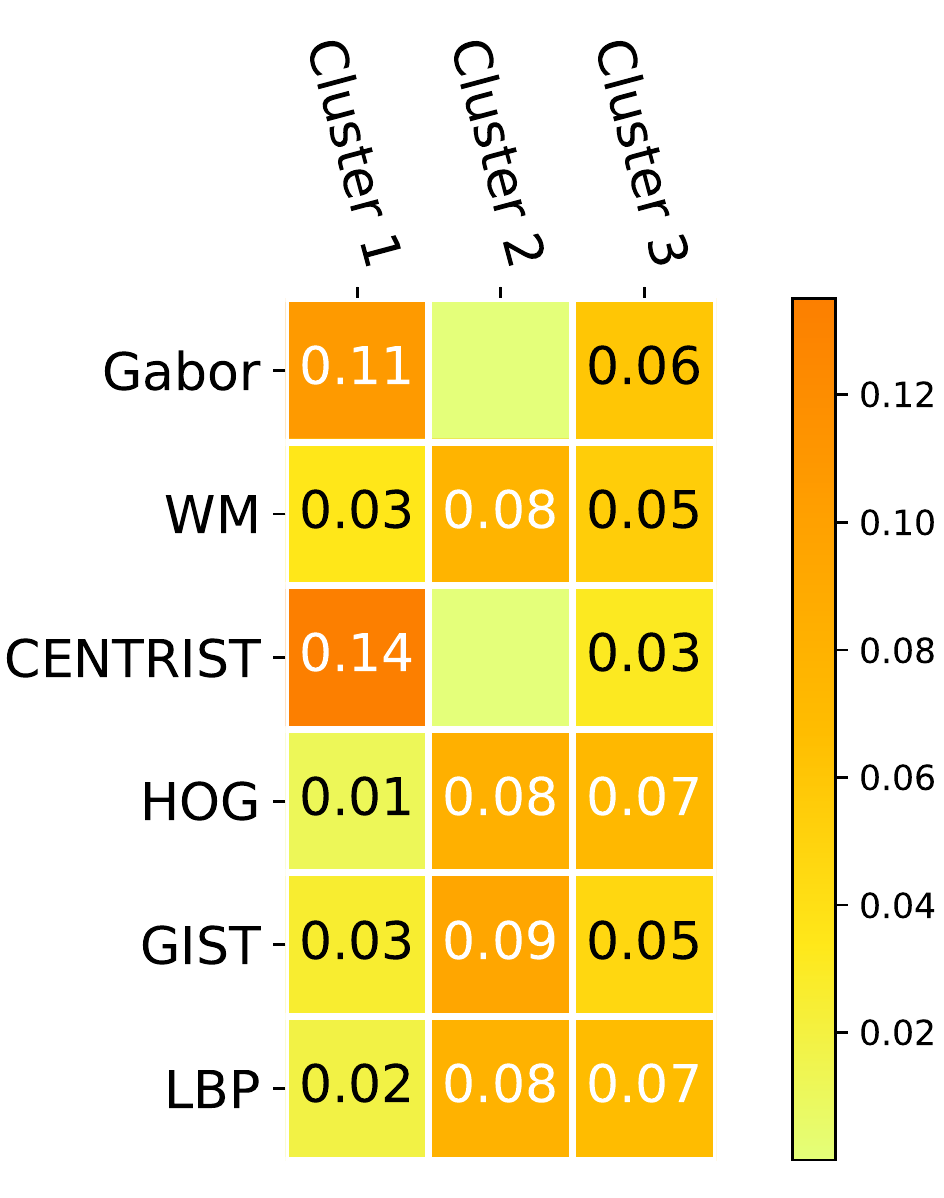}
    \includegraphics[height=3.5cm]{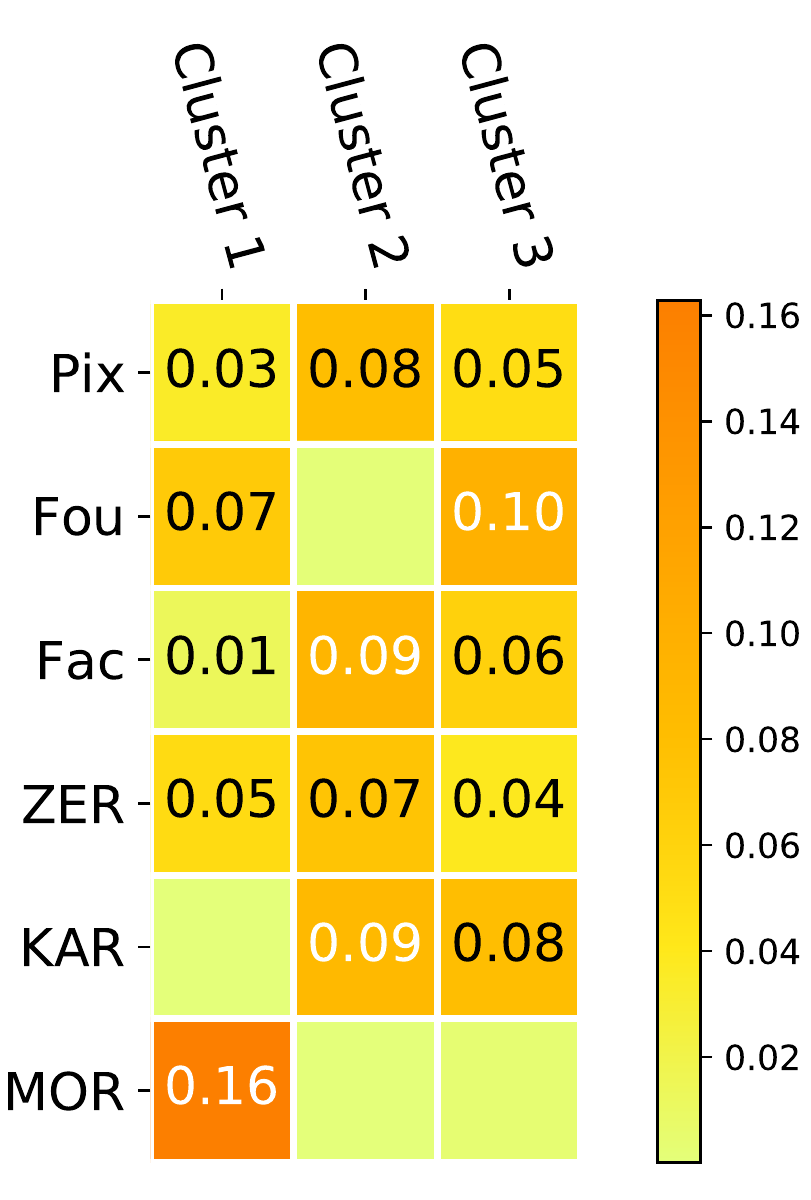}
    \vspace{-5pt}
    \caption{\small{Visualizations of the optimal transport matrices for the Caltech7 (left) and the Handwritten (right).
    }}
    \label{fig:ot}
\end{minipage}
~~
\begin{minipage}{0.56\textwidth}
\centering
\captionsetup{type=table} 
\begin{small}
    \centering
    \caption{Classification accuracy (\%) after removing some views from the datasets}\label{tab:result3}
    \vspace{-5pt}
    \begin{threeparttable}
    \begin{tabular}{
    @{\hspace{3pt}}c@{\hspace{3pt}}
    @{\hspace{3pt}}c@{\hspace{3pt}}|
    @{\hspace{3pt}}c@{\hspace{3pt}}
    @{\hspace{3pt}}c@{\hspace{3pt}}
    }
    \hline\hline
    Caltech7 &HOT (\ref{eq:sw-hot2}) &Handwritten &HOT (\ref{eq:sw-hot2}) \\ \hline
    All               &86.27$\pm$1.79 &All          &87.12$\pm$2.10\\
    \cancel{Gabor}    &86.03$\pm$1.90 &\cancel{Pix} &84.45$\pm$3.17\\
    \cancel{WM}       &86.15$\pm$1.69 &\cancel{Fou} &86.45$\pm$3.08\\
    \cancel{CENTRIST} &86.01$\pm$1.21 &\cancel{FAC} &84.65$\pm$2.57\\
    \cancel{HOG}      &82.95$\pm$2.12 &\cancel{ZER} &84.40$\pm$2.15\\
    \cancel{GIST}     &85.87$\pm$1.10 &\cancel{KAR} &84.75$\pm$3.38\\
    \cancel{LBP}      &85.43$\pm$2.84 &\cancel{MOR} &79.30$\pm$2.58\\
    \hline\hline
    \end{tabular}
    \begin{tablenotes}
    \item[1] ``All'' means training models using all the views.
    \item[2] ``\cancel{View}'' means removing the data of the view and training models accoridngly.
    \end{tablenotes}
    \end{threeparttable}
\end{small}
\end{minipage}
\end{figure}

\subsection{Comparisons on semi-supervised and unsupervised learning}
Applying the semi-supervised framework proposed in (\ref{eq:ssl-cca}), we test different multi-view learning methods and record their classification accuracy. 
For each method, we train $S$ multi-layer perceptron (MLP) models as encoders and a softmax layer as a classifier. 
To achieve fairness in these comparisons, all methods apply models with the same architecture and the same hyperparameters. 
In particular, we set the hyperparameters empirically as follows: the number of epochs is $100$; the learning rate is fixed as $0.001$; the batch size is $400$; for each $f_s$, the dimension of its output is $20$; the dimension of the common latent space is $10$; in (\ref{eq:ssl-cca}), $\tau=0.01$ and $\gamma=0.1$; in (\ref{eq:sw-hot1},~\ref{eq:sw-hot2}), $\alpha=0.01$; for the Sinkhorn algorithm, the number of iterations is $20$ and $\beta=0.1$; 
for sliced Wasserstein distance, the number of projections $M$ is set to be $3$ in our methods.
The robustness of our learning algorithms to the key hyperparameters above can be found in the Supplementary Material. 
The average performance of these methods for the 20 trials is reported in Table~\ref{tab:result1}. 
The baselines include the LSCCA \cite{via2007learning}, the DGCCA \cite{benton2019deep}, and the autoencoder-assisted CCA (AECCA) \cite{wang2015deep}. 
Compared with these baselines, which require well-aligned training data, our methods and their variants apply unaligned training data but achieve at least comparable performance on classification accuracy. 
Moreover, among our methods, applying our hierarchical optimal transport model generally achieves higher accuracy than applying sliced Wasserstein distance directly. 
These phenomena demonstrate the feasibility of sliced Wasserstein distance in multi-view learning and the advantage of our HOT model. 
Additionally, we can introduce a set of decoders corresponding to the encoders, and construct the regularizer $R_S$ as the reconstruction loss of the autoencoder (AE), which helps further improve the classification accuracy.

Besides semi-supervised learning, we can learn the latent representations of different views in an unsupervised manner, and then train a classifier based on them. 
The performance of different methods is shown in Table~\ref{tab:result2}, which helps us evaluate the power of different methods on unsupervised feature extraction. 
Compared with the performance achieved by semi-supervised learning, the performance of the baselines drops precipitously in this setting. 
For the proposed methods, those depending on pairwise comparisons between views ($i.e.$, implementing (\ref{eq:sw-p}, \ref{eq:sw-hot1})) suffer from the degradation of performance as well, while those learning clusters of views explicitly ($i.e.$, implementing (\ref{eq:sw-b}, \ref{eq:sw-hot2})) retain high classification accuracy, which implies that learning the clustering structure of views explicitly might be more suitable for unsupervised multi-view learning. 
Again, in this experiment applying our hierarchical optimal transport model can improve the performance.

\subsection{Justification of view clustering}
According to Table~\ref{tab:result2}, we find that applying our HOT model in (\ref{eq:sw-hot2}) achieves encouraging performance, which implies that the clustering structure of views learned by this method is reasonable. 
In Figure~\ref{fig:ot}, we visualize the corresponding optimal transport matrices learned for different datasets. 
We can find that for the views in the Caltech7 dataset, ``Gabor'' and  ``CENTIST'' belong to one cluster while ``WM'' and ``GIST'' belong to another, and ``HOG'' and ``LBP'' corresponds to the mixture of the cluster 2 and 3. 
To verify the rationality of this clustering structure, we evaluate the significance of different views in our learning tasks by removing each of the views and training the model accordingly. 
As shown in Table~\ref{tab:result3}, compared with the result achieved by using all views, removing either ``Gabor'' or ``CENTRIST'' (either ``WM'' or ``GIST'') just degrades the classification accuracy slightly, while removing ``HOG'' or ``LBP'' does harm to the accuracy severely. 
This result demonstrates that the clusters we find indeed group the views with redundant information and comparable contributions. 
Similarly, for the views in the Handwritten dataset, the distribution of ``Pix'' on the three clusters is similar to those of ``Fac'' and  ``KAR''. 
Therefore, removing one of them from the training views leads to similar classification accuracy. 
On the other hand, ``MOR'' is a unique view belonging to the cluster 1, so removing it leads to serious degradation on the performance.

\section{Conclusion}
We have proposed a hierarchical optimal transport model to achieve robust multi-view learning. 
This method neither depends on the correspondence between the samples of different views nor requires the views to obey the same latent distribution. 
The proposed approach consistently outperforms many strong baseline models on multiple datasets, demonstrating its potential for complicated learning tasks in real-world scenarios. 
In the future, we plan to explore the practical applications of our method, $e.g.$, introducing it to federated learning and achieving some predictive tasks for financial and healthcare data analysis. 
Additionally, we would like to extend our method to multi-view multi-task learning.

\newpage
\section{Broader Impact Statement}
This paper makes a significant contribution to extending the frontier of multi-view learning with fewer assumptions and better interpretability. 
To the best of our knowledge, our method is the first approach to learn latent representations for different views without correspondence and explore the clustering structure of the views at the same time. 
Its relationship to traditional multi-view learning methods is clarified in the paper as well. 
A potential application scenario of our work is multi-view learning based on distributed private data, $e.g.$, patients' records in different hospitals, and individuals' financial statements in different banks. 
Our method provides a potential solution to allow different organizations to share their data with better protections on data privacy: $i$) the data can be not only anonymous but also from different individuals; $ii$) in the training phase, we don't need to learn their correspondence explicitly. 
These advantages greatly reduce the risk of information leaking.

\bibliography{mvl_hot}

\begin{thebibliography}{10}\itemsep=-1pt

\bibitem{agueh2011barycenters}
M.~Agueh and G.~Carlier.
\newblock Barycenters in the wasserstein space.
\newblock {\em SIAM Journal on Mathematical Analysis}, 43(2):904--924, 2011.

\bibitem{andrew2013deep}
G.~Andrew, R.~Arora, J.~Bilmes, and K.~Livescu.
\newblock Deep canonical correlation analysis.
\newblock In {\em International conference on machine learning}, pages
  1247--1255, 2013.

\bibitem{arjovsky2017wasserstein}
M.~Arjovsky, S.~Chintala, and L.~Bottou.
\newblock Wasserstein gan.
\newblock {\em arXiv preprint arXiv:1701.07875}, 2017.

\bibitem{benamou2015iterative}
J.-D. Benamou, G.~Carlier, M.~Cuturi, L.~Nenna, and G.~Peyr{\'e}.
\newblock Iterative bregman projections for regularized transportation
  problems.
\newblock {\em SIAM Journal on Scientific Computing}, 37(2):A1111--A1138, 2015.

\bibitem{benton2019deep}
A.~Benton, H.~Khayrallah, B.~Gujral, D.~A. Reisinger, S.~Zhang, and R.~Arora.
\newblock Deep generalized canonical correlation analysis.
\newblock {\em ACL 2019}, page~1, 2019.

\bibitem{bonneel2015sliced}
N.~Bonneel, J.~Rabin, G.~Peyr{\'e}, and H.~Pfister.
\newblock Sliced and radon wasserstein barycenters of measures.
\newblock {\em Journal of Mathematical Imaging and Vision}, 51(1):22--45, 2015.

\bibitem{chaudhuri2009multi}
K.~Chaudhuri, S.~M. Kakade, K.~Livescu, and K.~Sridharan.
\newblock Multi-view clustering via canonical correlation analysis.
\newblock In {\em Proceedings of the 26th annual international conference on
  machine learning}, pages 129--136, 2009.

\bibitem{chen2018optimal}
Y.~Chen, T.~T. Georgiou, and A.~Tannenbaum.
\newblock Optimal transport for gaussian mixture models.
\newblock {\em IEEE Access}, 7:6269--6278, 2018.

\bibitem{christoudias2008multi}
C.~M. Christoudias, R.~Urtasun, and T.~Darrell.
\newblock Multi-view learning in the presence of view disagreement.
\newblock In {\em Proceedings of the Twenty-Fourth Conference on Uncertainty in
  Artificial Intelligence}, pages 88--96, 2008.

\bibitem{cuturi2013sinkhorn}
M.~Cuturi.
\newblock Sinkhorn distances: Lightspeed computation of optimal transport.
\newblock In {\em Advances in neural information processing systems}, pages
  2292--2300, 2013.

\bibitem{de2010multi}
V.~R. De~Sa, P.~W. Gallagher, J.~M. Lewis, and V.~L. Malave.
\newblock Multi-view kernel construction.
\newblock {\em Machine learning}, 79(1-2):47--71, 2010.

\bibitem{ding2014low}
Z.~Ding and Y.~Fu.
\newblock Low-rank common subspace for multi-view learning.
\newblock In {\em 2014 IEEE international conference on Data Mining}, pages
  110--119. IEEE, 2014.

\bibitem{guo2019canonical}
C.~Guo and D.~Wu.
\newblock Canonical correlation analysis (cca) based multi-view learning: An
  overview.
\newblock {\em arXiv preprint arXiv:1907.01693}, 2019.

\bibitem{guo2012cross}
Y.~Guo and M.~Xiao.
\newblock Cross language text classification via subspace co-regularized
  multi-view learning.
\newblock In {\em ICML}, 2012.

\bibitem{huang2018multimodal}
F.~Huang, X.~Zhang, C.~Li, Z.~Li, Y.~He, and Z.~Zhao.
\newblock Multimodal network embedding via attention based multi-view
  variational autoencoder.
\newblock In {\em Proceedings of the 2018 ACM on International Conference on
  Multimedia Retrieval}, pages 108--116, 2018.

\bibitem{jin2014multi}
X.~Jin, F.~Zhuang, H.~Xiong, C.~Du, P.~Luo, and Q.~He.
\newblock Multi-task multi-view learning for heterogeneous tasks.
\newblock In {\em Proceedings of the 23rd ACM international conference on
  conference on information and knowledge management}, pages 441--450, 2014.

\bibitem{kingma2014adam}
D.~P. Kingma and J.~Ba.
\newblock Adam: A method for stochastic optimization.
\newblock {\em arXiv preprint arXiv:1412.6980}, 2014.

\bibitem{kolouri2018sliced}
S.~Kolouri, G.~K. Rohde, and H.~Hoffmann.
\newblock Sliced wasserstein distance for learning gaussian mixture models.
\newblock In {\em Proceedings of the IEEE Conference on Computer Vision and
  Pattern Recognition}, pages 3427--3436, 2018.

\bibitem{kumar2011co}
A.~Kumar and H.~Daum{\'e}.
\newblock A co-training approach for multi-view spectral clustering.
\newblock In {\em Proceedings of the 28th international conference on machine
  learning (ICML-11)}, pages 393--400, 2011.

\bibitem{kusner2015word}
M.~Kusner, Y.~Sun, N.~Kolkin, and K.~Weinberger.
\newblock From word embeddings to document distances.
\newblock In {\em International conference on machine learning}, pages
  957--966, 2015.

\bibitem{lee2019hierarchical}
J.~Lee, M.~Dabagia, E.~Dyer, and C.~Rozell.
\newblock Hierarchical optimal transport for multimodal distribution alignment.
\newblock In {\em Advances in Neural Information Processing Systems}, pages
  13453--13463, 2019.

\bibitem{li2015large}
Y.~Li, F.~Nie, H.~Huang, and J.~Huang.
\newblock Large-scale multi-view spectral clustering via bipartite graph.
\newblock In {\em Twenty-Ninth AAAI Conference on Artificial Intelligence},
  2015.

\bibitem{li2018survey}
Y.~Li, M.~Yang, and Z.~Zhang.
\newblock A survey of multi-view representation learning.
\newblock {\em IEEE transactions on knowledge and data engineering},
  31(10):1863--1883, 2018.

\bibitem{ma2017self}
F.~Ma, D.~Meng, Q.~Xie, Z.~Li, and X.~Dong.
\newblock Self-paced co-training.
\newblock In {\em Proceedings of the 34th International Conference on Machine
  Learning-Volume 70}, pages 2275--2284. JMLR. org, 2017.

\bibitem{memoli2011gromov}
F.~M{\'e}moli.
\newblock Gromov--wasserstein distances and the metric approach to object
  matching.
\newblock {\em Foundations of computational mathematics}, 11(4):417--487, 2011.

\bibitem{quang2013unifying}
M.~H. Quang, L.~Bazzani, and V.~Murino.
\newblock A unifying framework for vector-valued manifold regularization and
  multi-view learning.
\newblock In {\em International conference on machine learning}, pages
  100--108, 2013.

\bibitem{schmitzer2013hierarchical}
B.~Schmitzer and C.~Schn{\"o}rr.
\newblock A hierarchical approach to optimal transport.
\newblock In {\em International Conference on Scale Space and Variational
  Methods in Computer Vision}, pages 452--464. Springer, 2013.

\bibitem{sindhwani2008rkhs}
V.~Sindhwani and D.~S. Rosenberg.
\newblock An rkhs for multi-view learning and manifold co-regularization.
\newblock In {\em Proceedings of the 25th international conference on Machine
  learning}, pages 976--983, 2008.

\bibitem{su2017order}
B.~Su and G.~Hua.
\newblock Order-preserving wasserstein distance for sequence matching.
\newblock In {\em Proceedings of the IEEE Conference on Computer Vision and
  Pattern Recognition}, pages 1049--1057, 2017.

\bibitem{sun2013survey}
S.~Sun.
\newblock A survey of multi-view machine learning.
\newblock {\em Neural computing and applications}, 23(7-8):2031--2038, 2013.

\bibitem{via2007learning}
J.~V{\'\i}a, I.~Santamar{\'\i}a, and J.~P{\'e}rez.
\newblock A learning algorithm for adaptive canonical correlation analysis of
  several data sets.
\newblock {\em Neural Networks}, 20(1):139--152, 2007.

\bibitem{villani2008optimal}
C.~Villani.
\newblock {\em Optimal transport: old and new}, volume 338.
\newblock Springer Science \& Business Media, 2008.

\bibitem{wang2015deep}
W.~Wang, R.~Arora, K.~Livescu, and J.~Bilmes.
\newblock On deep multi-view representation learning.
\newblock In {\em International Conference on Machine Learning}, pages
  1083--1092, 2015.

\bibitem{wang2019adversarial}
X.~Wang, D.~Peng, P.~Hu, and Y.~Sang.
\newblock Adversarial correlated autoencoder for unsupervised multi-view
  representation learning.
\newblock {\em Knowledge-Based Systems}, 168:109--120, 2019.

\bibitem{white2012convex}
M.~White, X.~Zhang, D.~Schuurmans, and Y.-l. Yu.
\newblock Convex multi-view subspace learning.
\newblock In {\em Advances in Neural Information Processing Systems}, pages
  1673--1681, 2012.

\bibitem{xu2015multi}
C.~Xu, D.~Tao, and C.~Xu.
\newblock Multi-view learning with incomplete views.
\newblock {\em IEEE Transactions on Image Processing}, 24(12):5812--5825, 2015.

\bibitem{xu2020gromov}
H.~Xu.
\newblock Gromov-wasserstein factorization models for graph clustering.
\newblock {\em arXiv preprint arXiv:1911.08530}, 2019.

\bibitem{xu2020learning}
H.~Xu, D.~Luo, R.~Henao, S.~Shah, and L.~Carin.
\newblock Learning autoencoders with relational regularization.
\newblock {\em arXiv preprint arXiv:2002.02913}, 2020.

\bibitem{xu2018distilled}
H.~Xu, W.~Wang, W.~Liu, and L.~Carin.
\newblock Distilled wasserstein learning for word embedding and topic modeling.
\newblock In {\em Advances in Neural Information Processing Systems}, pages
  1716--1725, 2018.

\bibitem{ye2016learning}
T.~Ye, T.~Wang, K.~McGuinness, Y.~Guo, and C.~Gurrin.
\newblock Learning multiple views with orthogonal denoising autoencoders.
\newblock In {\em International Conference on Multimedia Modeling}, pages
  313--324. Springer, 2016.

\bibitem{yuan2018multi}
Y.~Yuan, G.~Xun, K.~Jia, and A.~Zhang.
\newblock A multi-view deep learning framework for eeg seizure detection.
\newblock {\em IEEE journal of biomedical and health informatics},
  23(1):83--94, 2018.

\bibitem{yurochkin2019hierarchical}
M.~Yurochkin, S.~Claici, E.~Chien, F.~Mirzazadeh, and J.~M. Solomon.
\newblock Hierarchical optimal transport for document representation.
\newblock In {\em Advances in Neural Information Processing Systems}, pages
  1599--1609, 2019.

\bibitem{zhang2018multi}
C.~Zhang, E.~Adeli, T.~Zhou, X.~Chen, and D.~Shen.
\newblock Multi-layer multi-view classification for alzheimer’s disease
  diagnosis.
\newblock In {\em Thirty-Second AAAI Conference on Artificial Intelligence},
  2018.

\bibitem{zhao2017multi}
J.~Zhao, X.~Xie, X.~Xu, and S.~Sun.
\newblock Multi-view learning overview: Recent progress and new challenges.
\newblock {\em Information Fusion}, 38:43--54, 2017.

\end{thebibliography}
\bibliographystyle{ieee}

\newpage
\section{Supplementary Material}
\subsection{The proof of Proposition~\ref{prop1}}\label{app:1}
\textbf{Proposition~\ref{prop1}} \textit{Given two sets of samples, denoted as $Z_1$ and $Z_2$, each of which has $N$ $d$-dimensional samples, $\min_{P\in\mathcal{P}}\|Z_1P-Z_2\|_F^2\geq \widehat{D}_{\text{sw}}(Z_1, Z_2)$.}
\begin{proof}
Given two sets of samples, denoted as $Z_1$ and $Z_2$, each of which has $N$ $d$-dimensional samples, we have 
\begin{eqnarray}
\begin{aligned}
\widehat{D}_{\text{sw}}(Z_1, Z_2) &= \frac{1}{M}\sideset{}{_{m=1}^{M}}\sum\sideset{}{_{P\in\mathcal{P}}}\min\|\theta_m^{\top}Z_1P-\theta_m^{\top}Z_2\|_2^2\\
&\leq \frac{1}{M}\sideset{}{_{P\in\mathcal{P}}}\min\sideset{}{_{m=1}^{M}}\sum\|\theta_m^{\top}Z_1P-\theta_m^{\top}Z_2\|_2^2\\
&= \frac{1}{M}\sideset{}{_{P\in\mathcal{P}}}\min\|\Theta^{\top} Z_1 P - \Theta^{\top} Z_2\|_F^2\\
&\leq \frac{1}{M}\sideset{}{_{P\in\mathcal{P}}}\min\|\Theta\|_F^2\| Z_1 P -  Z_2\|_F^2\\
&=\sideset{}{_{P\in\mathcal{P}}}\min\| Z_1 P -  Z_2\|_F^2.
\end{aligned}
\end{eqnarray}
Here, $\Theta=[\theta_1,...,\theta_M]$. 
Because each $\theta_m\in\mathcal{S}^{d-1}$, we have $\|\Theta\|_F^2=M$.
\end{proof}

\subsection{The Sinkhorn scaling algorithm}\label{app:2}
The scheme of the Sinkhorn scaling algorithm is shown below:
\begin{algorithm}[h]
\small{
    \caption{$\min_{W\in \Pi(p,q)}\langle W, C \rangle+\beta\langle W, \log W \rangle$}	
    \label{alg:proximal}
	\begin{algorithmic}[1]
	    \STATE Initialize ${T}^{(0)}=ab^{\top}$, $a=p$
	    \STATE $\Phi=\exp(-\frac{1}{\beta}C)$.
		\STATE \textbf{for} $j=0,...,J-1$ 
		\STATE \quad Sinkhorn iteration: \\
		\quad\quad ${b}=\frac{q}{{\Phi}^{\top}{a}}$\\
		\quad\quad ${a} = \frac{{p}}{{\Phi}{b}}$
		\STATE \textbf{Return} ${T}^{(J)}=\text{diag}({a}){\Phi}\text{diag}({b})$
	\end{algorithmic}
}
\end{algorithm}

\subsection{The configuration of models and hyperparameters}
We implement all the models with PyTorch and train them on a single NVIDIA GTX 1080 Ti GPU.
For our methods, the hyperparameters are set empirically as follows: the number of epochs is $100$; the learning rate is fixed as $0.001$; the batch size is $400$; for each $f_s$, the dimension of its output is 20; the dimension of the common latent space is 10; in (\ref{eq:ssl-cca}), $\tau=0.01$ and $\gamma=0.1$; in (\ref{eq:sw-hot1},~\ref{eq:sw-hot2}), $\alpha=0.01$; for the Sinkhorn algorithm, the number of iterations is $20$ and $\beta=0.1$; 
for sliced Wasserstein distance, the number of projections $M$ is set to be $3$ in our methods; 
for the HOT in (\ref{eq:sw-hot2}), the number of the clusters of views $K$ is set to be $3$.

\begin{figure}[t]
    \centering
    \subfigure[Accuracy v.s. batch size]{
    \includegraphics[width=0.31\linewidth]{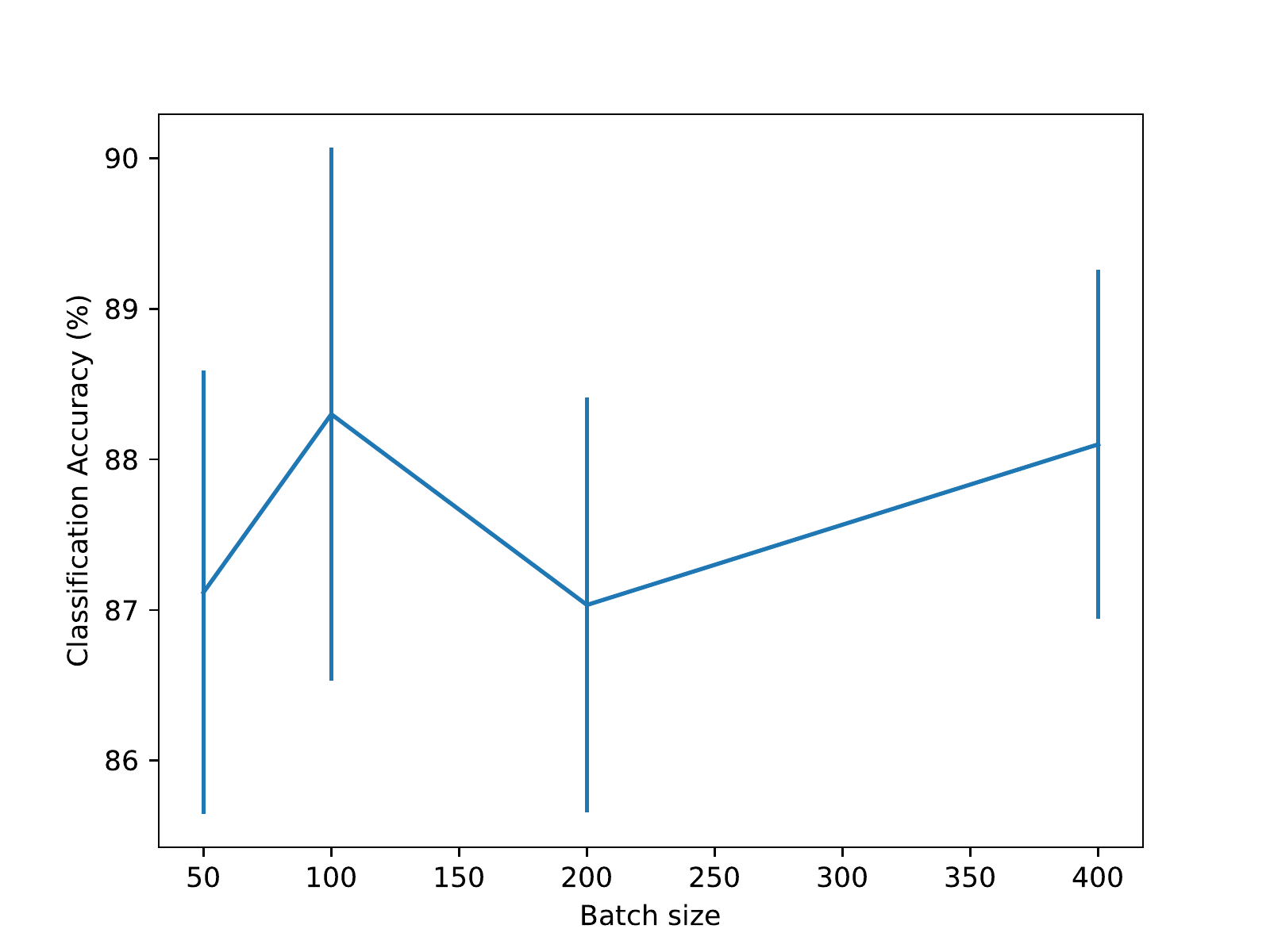}\label{fig:batch}
    }
    \subfigure[Accuracy v.s. \# projections]{
    \includegraphics[width=0.31\linewidth]{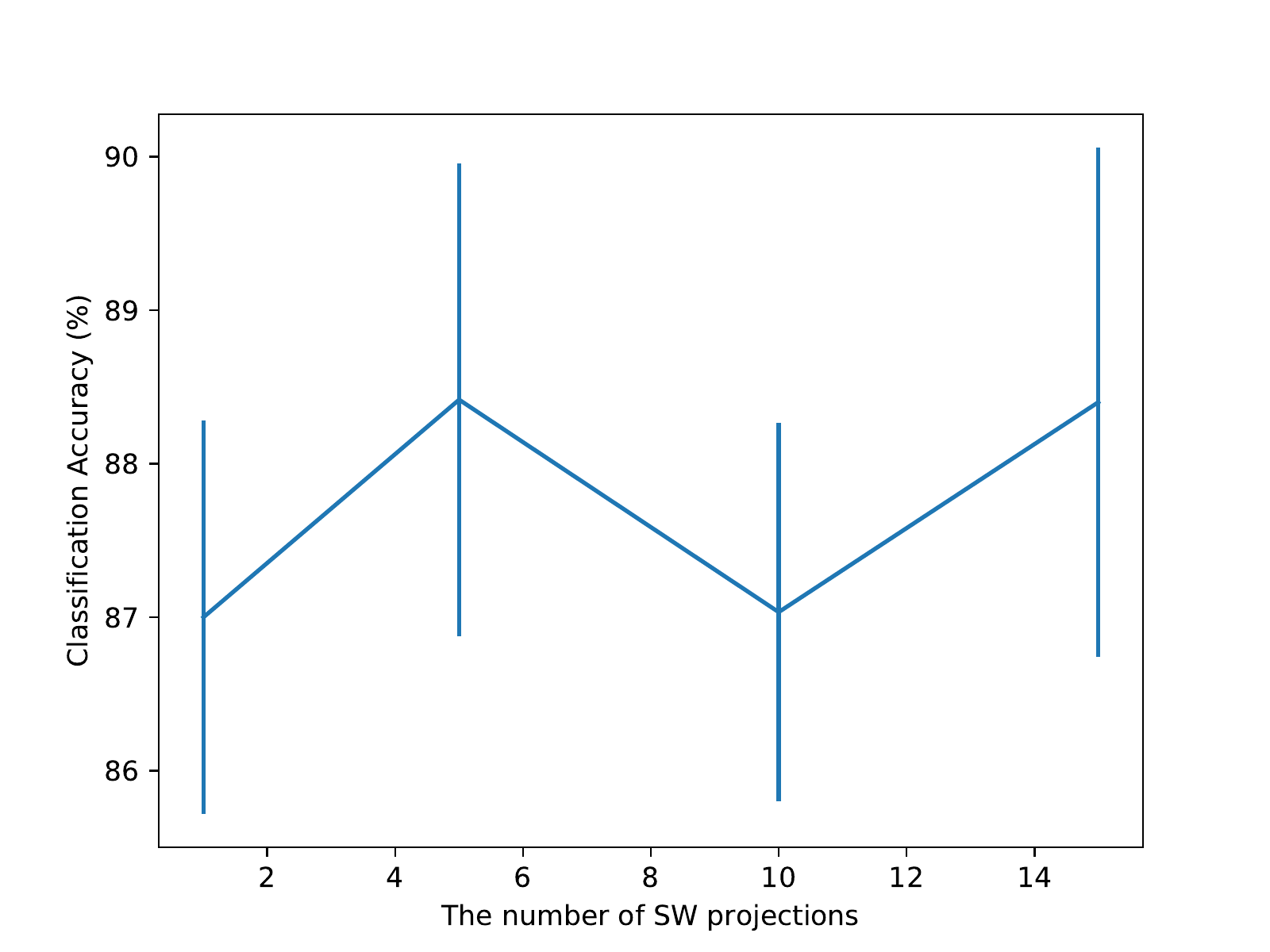}\label{fig:proj}
    }
    \subfigure[Accuracy v.s. \# clusters]{
    \includegraphics[width=0.31\linewidth]{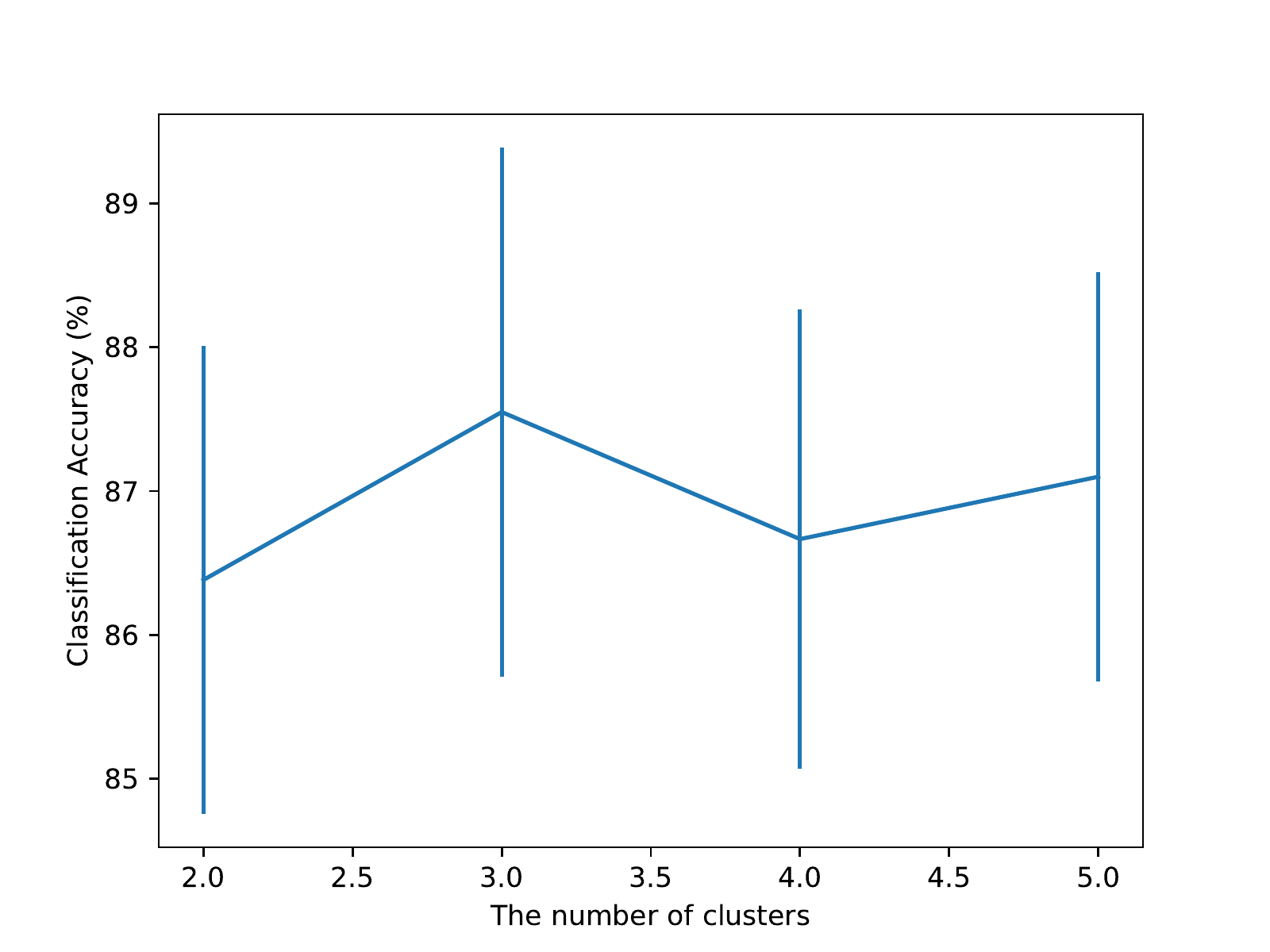}\label{fig:k}
    }
    \vspace{-5pt}
    \caption{\small{The robustness of our method (the HOT in (\ref{eq:sw-hot2})) to its hyperparameters in unsupervised learning. We use the Handwritten dataset in this experiment.}}
    \label{fig:robust}
\end{figure}

\begin{figure}[t]
    \centering
    \subfigure[Accuracy v.s. $\alpha$]{
    \includegraphics[width=0.31\linewidth]{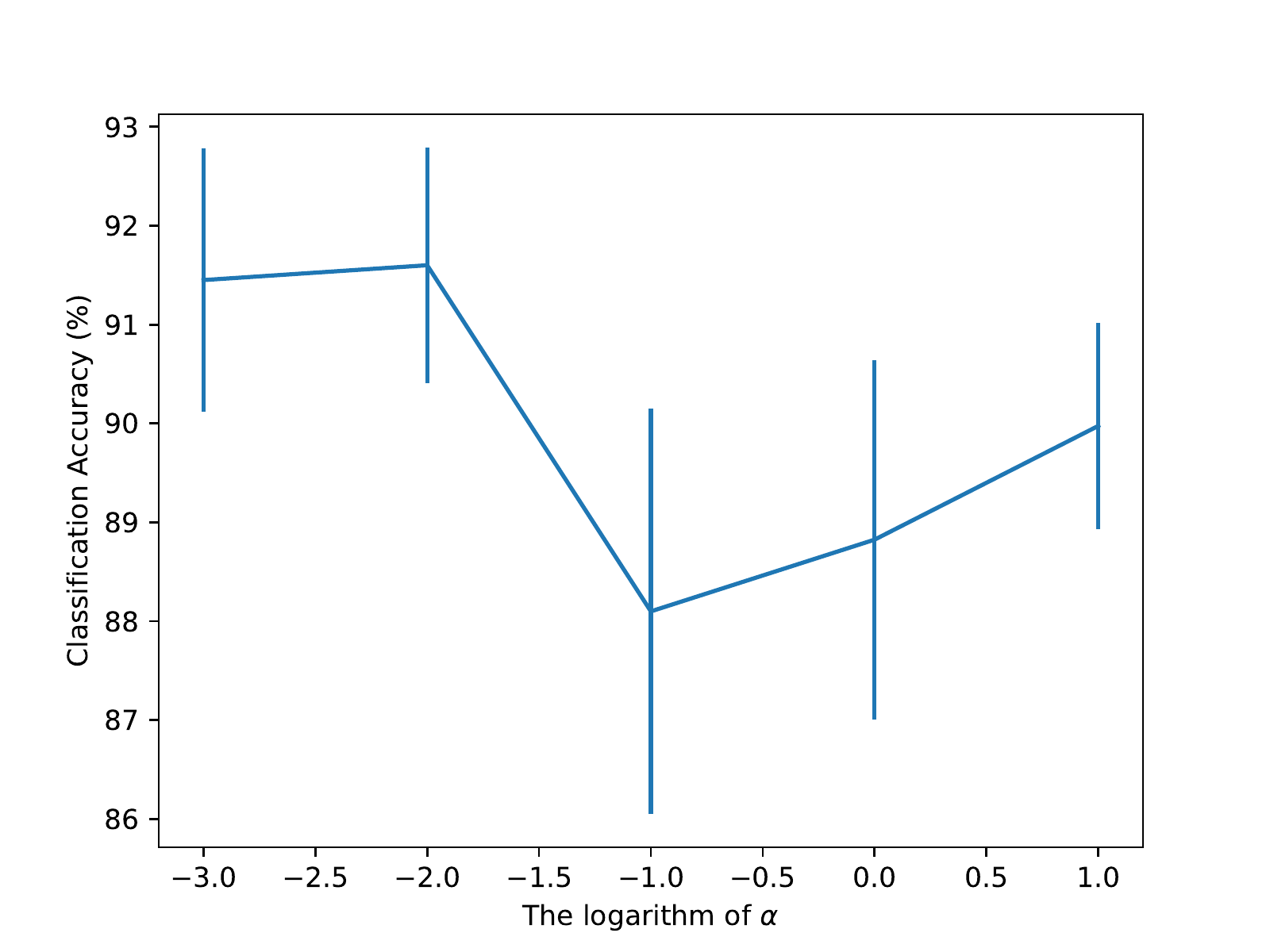}\label{fig:alpha}
    }
    \subfigure[Accuracy v.s. $\tau$]{
    \includegraphics[width=0.31\linewidth]{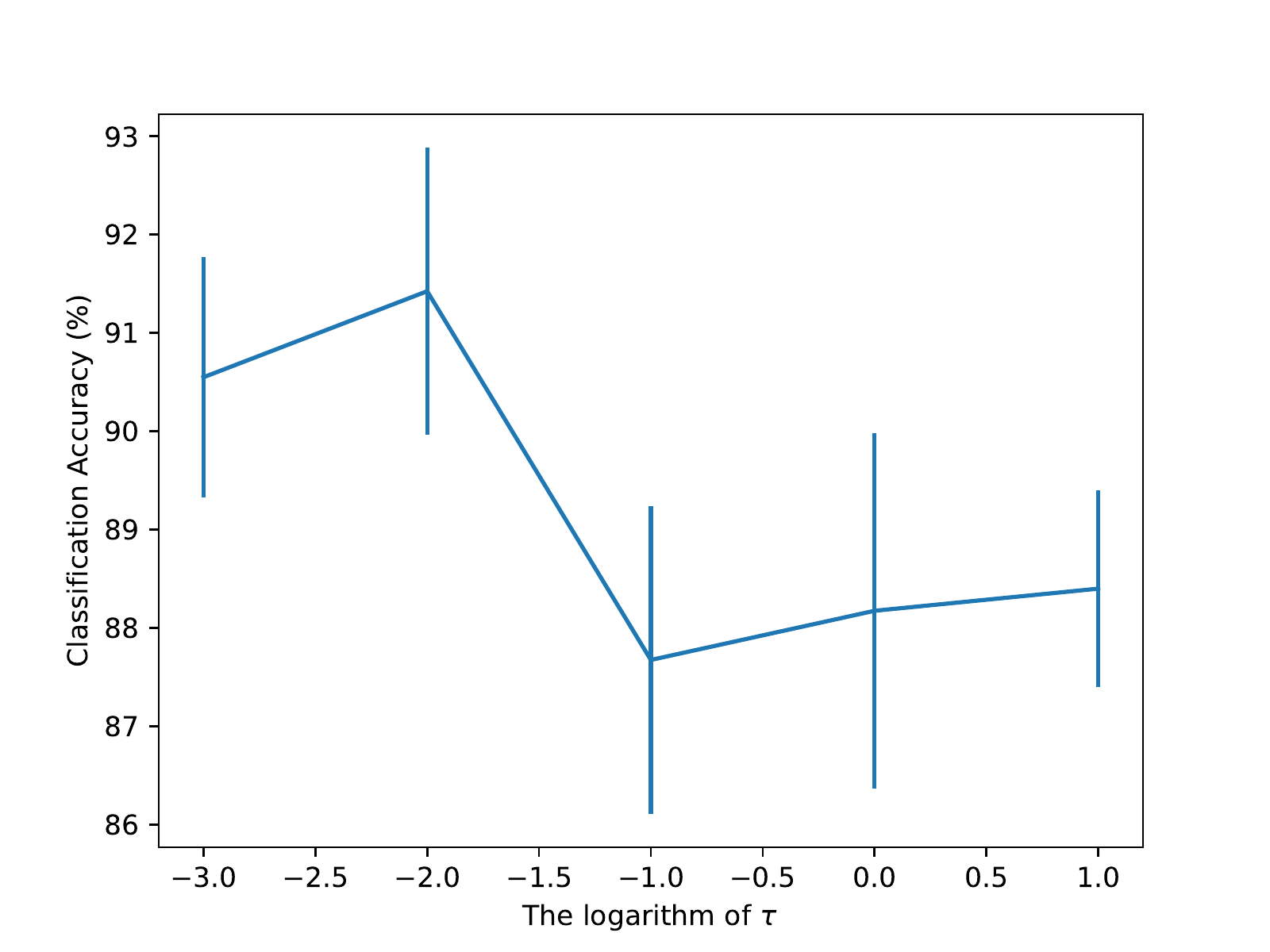}\label{fig:tau}
    }
    \vspace{-5pt}
    \caption{\small{The robustness of our method (the HOT in (\ref{eq:sw-hot2})) to its hyperparameters in semi-supervised learning. We use the Handwritten dataset in this experiment.}}
    \label{fig:robust2}
\end{figure}

Among these hyperparameters, there are three key hyperparameters: the batch size, the number of projections when calculating the sliced Wasserstein distance, and the number of clusters $K$. 
In particular, the sliced Wasserstein distance used in our work provides an empirical estimation for the expected distance between distributions based on their samples. 
The batch size controls the number of samples used to calculate the sliced Wasserstein distance. 
The number of projections controls the precision and the stability of the estimation. 
Generally, using a large batch size and a large number of projections provides us better estimation but increases computations at the same time.
In Figure~\ref{fig:batch} and Figure~\ref{fig:proj}, we can find that the performance of our multi-view learning method (the HOT in (\ref{eq:sw-hot2})) is relatively robust to the change of these two hyperparameters. 
According to these two figures, we set the batch size to be 400 and the number of projections be 3. 
Similarly, Figure~\ref{fig:k} shows that our method is robust to the number of clusters we set, and the best performance is achieved when $K=3$. 
For semi-supervised learning, there are two more key hyperparameters: the weight of the multi-view learning regularizer $\tau$ and the weight $\alpha$ in (\ref{eq:sw-hot1}, \ref{eq:sw-hot2}).
According to Figure~\ref{fig:robust2}, when $\alpha$ and $\tau$ are set to be $0.01$ our method can achieve encouraging performance.

\subsection{More results of optimal transport matrices}
\begin{figure}[t]
    \centering
    \subfigure[The Caltech7]{
    \includegraphics[width=0.36\linewidth]{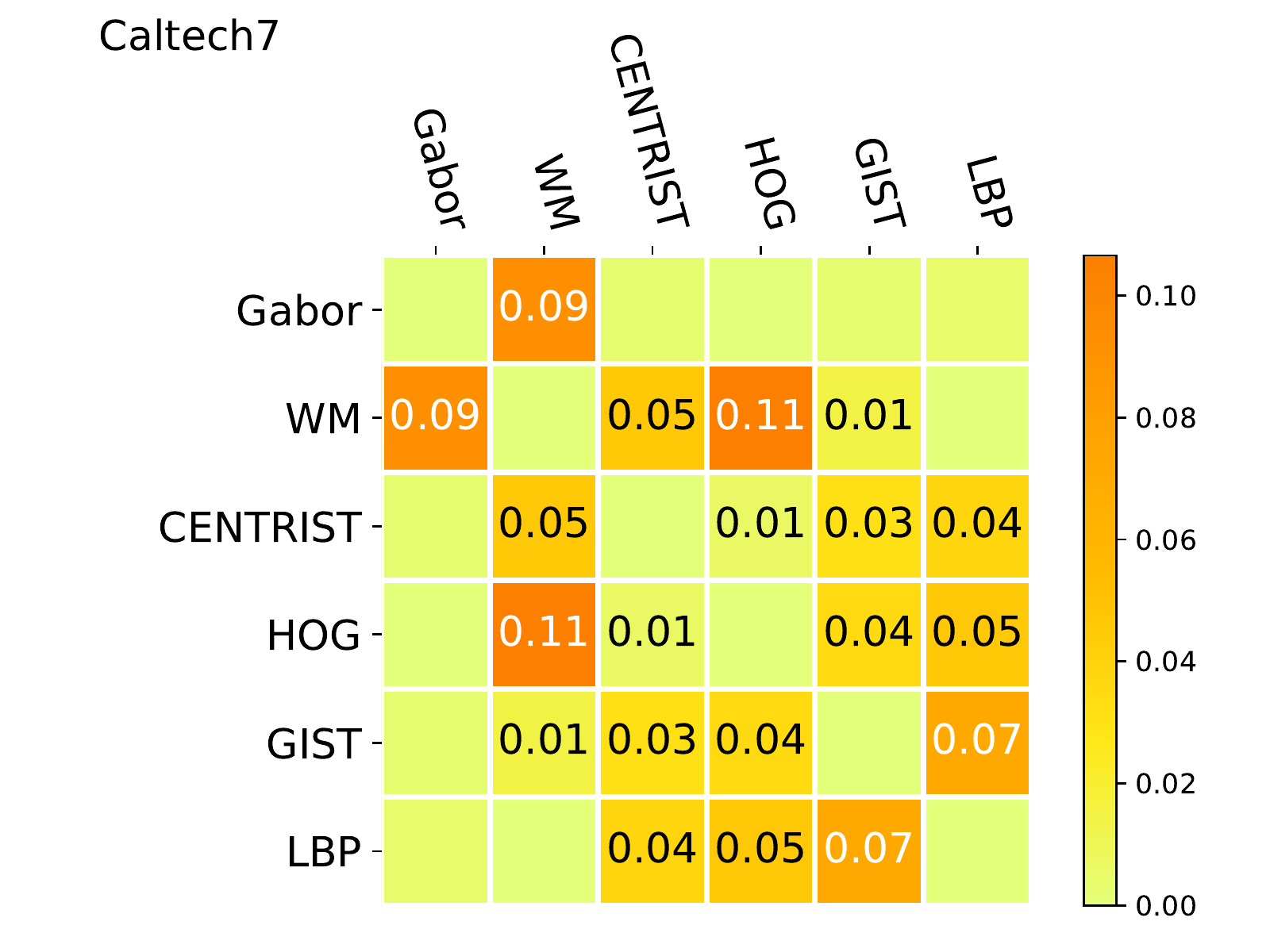}
    \includegraphics[width=0.36\linewidth]{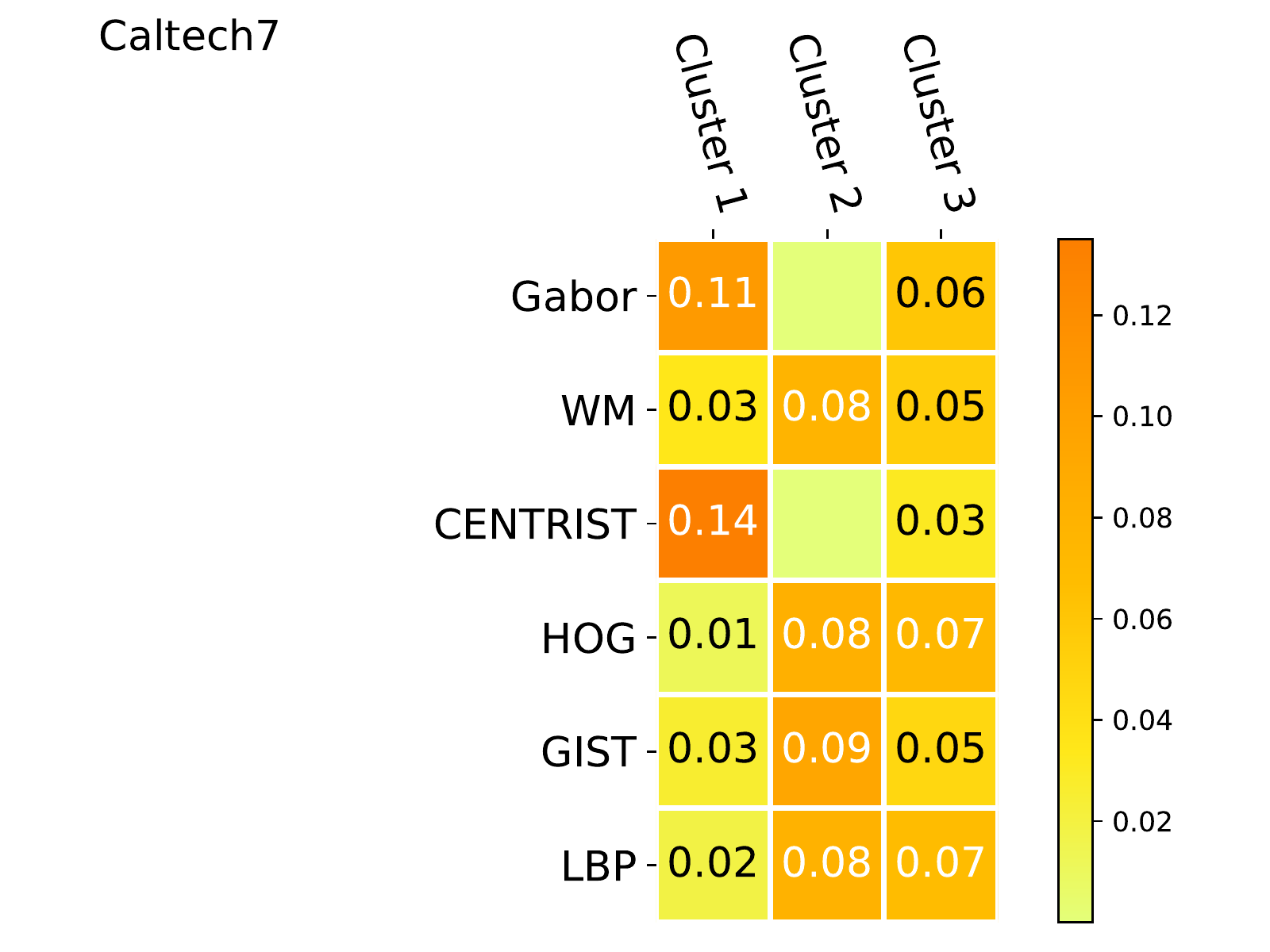}
    }\\
    \subfigure[The Caltech20]{
    \includegraphics[width=0.36\linewidth]{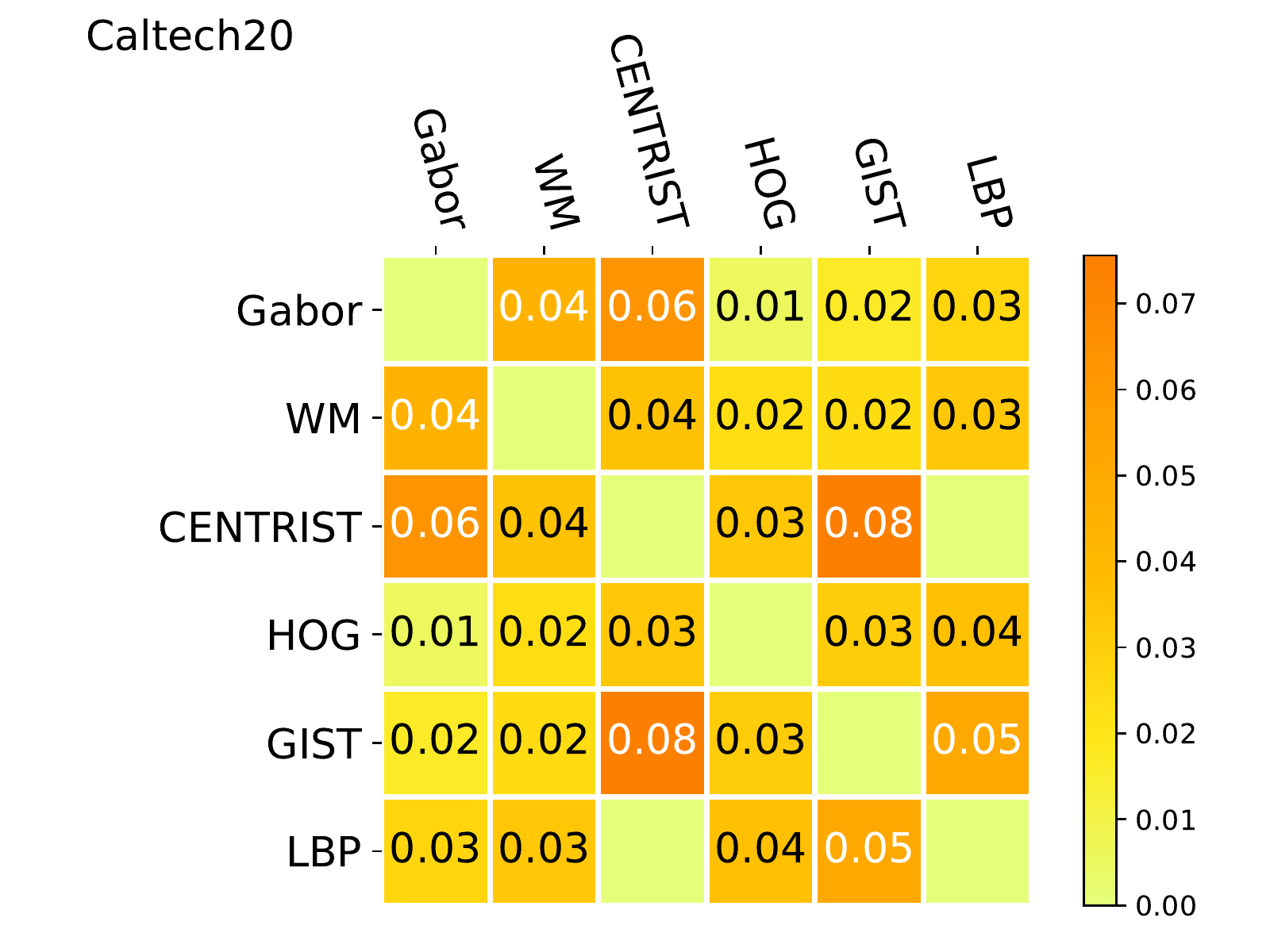}
    \includegraphics[width=0.36\linewidth]{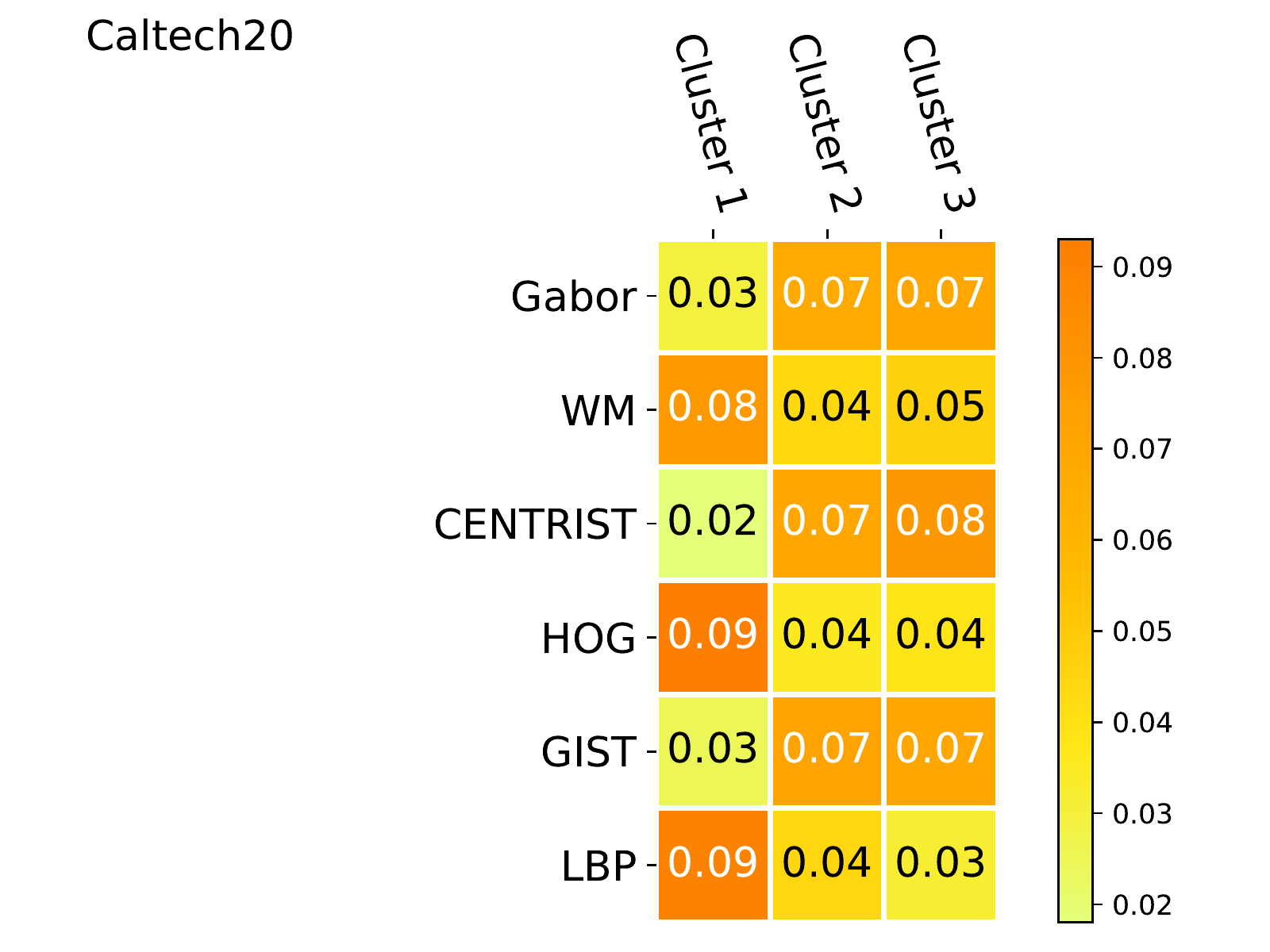}
    }\\
    \subfigure[The Handwritten]{
    \includegraphics[width=0.36\linewidth]{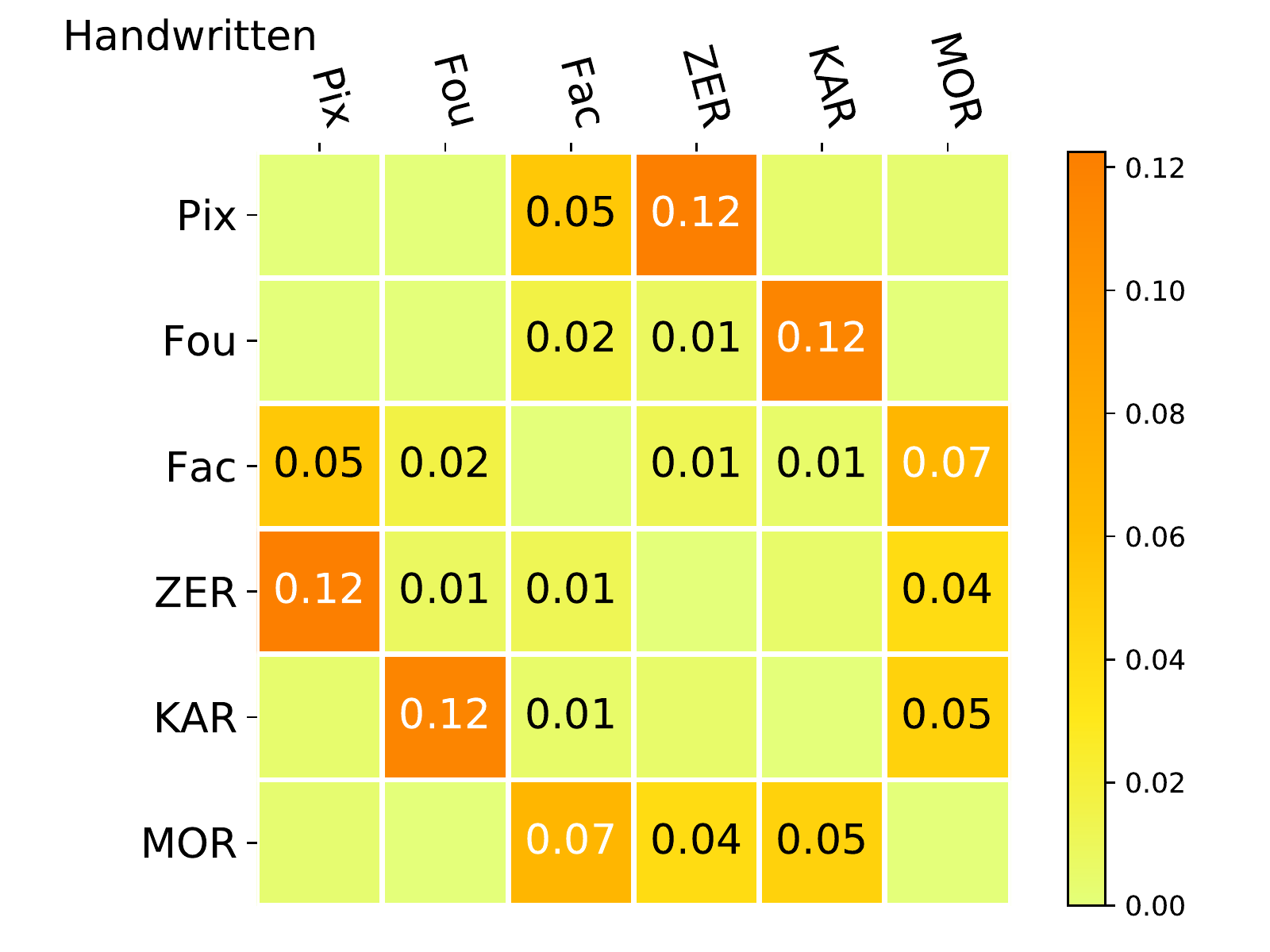}
    \includegraphics[width=0.36\linewidth]{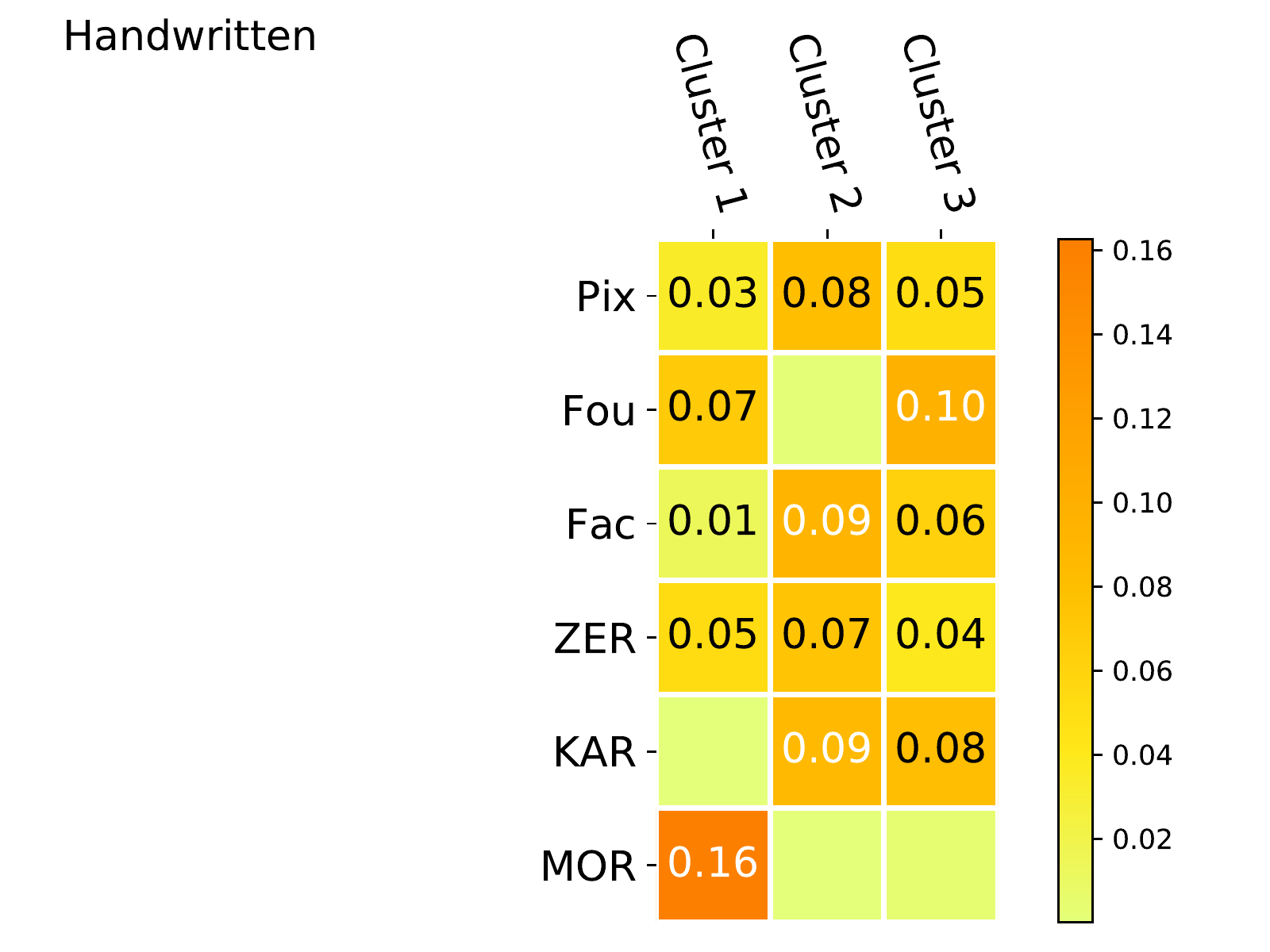}
    }
    \vspace{-5pt}
    \caption{\small{The visualizations of the optimal transport matrices for the views.
    In each subfigure, the left is the optimal transport matrix leaned by the HOT in (\ref{eq:sw-hot1}), and the right is that learned by solving (\ref{eq:sw-hot2}).}}
    \label{fig:moreot}
\end{figure}

Figure~\ref{fig:moreot} visualize the optimal transport matrices obtained for different datasets when we learn the latent representations of their views in unsupervised ways. 
We can find that both of our two HOT models can predict the clusters of the views. 
However, the clustering structures learned by them are inconsistent in some cases. 
For example, for the Caltech7 dataset, ``Gabor'', ``WM'', and ``HOG'' are likely to be in the same cluster when we apply the HOT model in~\ref{eq:sw-hot1}. 
On the other hand, when we use the HOT model in~\ref{eq:sw-hot2}, ``Gabor'' is more likely to be grouped with ``CENTRIST''. 
Because the performance of the second model is better than that of the first one for unsupervised learning, we think the clusters detected by the second model is more reliable, which is also verified in the Table~\ref{tab:result3} in the main paper. 

Additionally, we find that for the simple tasks ($e.g.$, the Caltech7 and the Handwritten), our methods can learn sparse optimal transport matrices and detect clusters clearly. 
For complicated tasks, $e.g.$, those with many classes (the Caltech20), the optimal transport matrices we learned are often dense, and the clusters are not so obvious as those in the simple cases.

\end{document}